\pdfoutput=1

\documentclass[11pt]{article}

\usepackage[preprint]{acl}

\usepackage{times}
\usepackage{latexsym}

\usepackage[T1]{fontenc}

\usepackage{microtype}

\usepackage{inconsolata}

\usepackage{graphicx}

\title{Cognitive Linguistic Identity Fusion Score (CLIFS): A Scalable Cognition‑Informed Approach to Quantifying Identity Fusion from Text}

\author{
 \textbf{Devin R. Wright\textsuperscript{1,2,4}},
 \textbf{Jisun An\textsuperscript{1}},
 \textbf{Yong-Yeol Ahn\textsuperscript{3}},
\\
\\
 \textsuperscript{1}Center for Complex Networks and Systems Research, Luddy School of Informatics,\\Computing, and Engineering, Indiana University Bloomington\\
 \textsuperscript{2}Cognitive Science Program, Indiana University Bloomington \\
 \textsuperscript{3}School of Data Science, University of Virginia \\
 \textsuperscript{4}CulturePulse, Inc.
\\
 \small{
   \href{mailto:devrwrig@iu.edu}{devrwrig@iu.edu}\quad\href{mailto:jisunan@iu.edu} {jisunan@iu.edu}\quad\href{mailto:yyahn@virginia.edu}{yyahn@virginia.edu}
 }
}

\begin{document}
\maketitle
\begin{abstract}

Quantifying \emph{identity fusion}---the psychological merging of self with another entity or abstract target (e.g., a religious group, political party, ideology, value, brand, belief, etc.)---is vital for understanding a wide range of group‑based human behaviors. We introduce the Cognitive Linguistic Identity Fusion Score (\href{https://github.com/DevinW-sudo/CLIFS}{CLIFS}), a novel metric that integrates cognitive linguistics with large language models (LLMs), which builds on implicit metaphor detection. Unlike traditional pictorial and verbal scales, which require controlled surveys or direct field contact, CLIFS delivers fully automated, scalable assessments while maintaining strong alignment with the established verbal measure. In benchmarks, CLIFS outperforms both existing automated approaches and human annotation. As a proof of concept, we apply CLIFS to violence risk assessment to demonstrate that it can improve violence risk assessment by more than 240\%. Building on our identification of a new NLP task and early success, we underscore the need to develop larger, more diverse datasets that encompass additional fusion-target domains and cultural backgrounds to enhance generalizability and further advance this emerging area. CLIFS models and code are public at \url{https://github.com/DevinW-sudo/CLIFS}.

\end{abstract}

\section{Introduction}
\renewcommand{\thefootnote}{\fnsymbol{footnote}}
\footnotetext[2]{\textbf{Correspondence:} \href{mailto:devrwrig@iu.edu}
{devrwrig@iu.edu}}
\renewcommand{\thefootnote}{\arabic{footnote}}
In Comprehensive Identity Fusion Theory (\textbf{\textsc{cift}})\footnote{For a helpful reference table of acronyms and symbols used or introduced in this paper, see Table \ref{tab:acronyms_symbols} in Appendix \ref{sec:appendixd}.}, identity fusion is commonly referred to as a ``visceral feeling of oneness,'' often felt by an individual with a group~\citep{SwannKleinGomez2024,SwannJettenGomezWhitehouseBastian2012, SwannGomezEtAl2009}. In contrast to the traditional social identity theory~\cite{TajfelTurner1979}, \textsc{cift} suggests that identity fusion is a unique form of group alignment that can occur not only with social groups but also with any abstract target such as an ideology, leader, value, or belief~\citep{SwannKleinGomez2024}. Identity fusion is a stable alignment where the personal self remains active and mutually reinforcing with the fusion target identity, characterized by porous boundaries and a tendency to motivate both extreme and prosocial in-group behavior~\citep{SwannKleinGomez2024}.

Identity fusion manifests itself in various ways; examples include extreme self-sacrifice and defense of the target group---e.g. fighting, killing, or dying for their target group, prioritizing fused target over family, and even support for honor violence or denial of in-group wrongdoing~\citep{SwannKleinGomez2024,AshokkumarSwann2023,BestaGomezVazquez2014,SwannBuhrmesterGomezEtAl2014,WhitehouseMcQuinnBuhrmesterSwann2014,SwannGomezDovidioHartJetten2010}. Fusion can also drive enacted or endorsed political persecution and violent opposition to unfavorable political outcomes~\citep{KunstDovidioThomsen2019}. Recent work reveals a more nuanced role: fusion correlates with social exploration and out-group trust in peaceful settings, suggesting it can support intergroup cooperation absent perceived existential threats~\citep{KleinGreenawayBastian2024}.

Although prior research has uncovered various pathways leading individuals toward and away (known as ``defusion'') from identity fusion, some defusion methods can be ethically problematic or even backfire (e.g., imprisonment, solitary confinement, degrading social support systems, seeding doubt and distrust of in-group), and there remain a lot of knowledge gaps in both fusion and defusion~\citep{SwannKleinGomez2024,GomezChinchillaVazquezEtAl2020}. The ability to quantitatively estimate identity fusion is important, given that it can drive powerful social consequences. These consequences manifest as beneficial outcomes---such as enhanced social cohesion and prosocial behaviors---and harmful outcomes---such as radicalization and violence. Advancing this line of inquiry requires tools to estimate the strength of fusion and reliably track it longitudinally across larger populations.

Despite advances in understanding identity fusion and its consequential nature for stable, cooperative, and cohesive social systems; empirical measures remain largely self-reported or qualitative~\citep{EbnerKavanaghWhitehouse2022,JimenezGomezEtAl2016,GomezBrooksEtAl2011,SwannGomezEtAl2009}. This gap precludes large-scale, longitudinal, and historical analyses of inter- and intra-community dynamics of identify fusion, the mechanisms
that shape fusion processes, and the spectrum of fusion outcomes, from destructive violence to social cohesion and cooperation.~\citep{EbnerKavanaghWhitehouse2022,KleinGreenawayBastian2024}.

Here, we introduce ``Cognitive Linguistic Identity Fusion Score (\textbf{CLIFS}),'' an automated, text-based metric of identity fusion that leverages LLMs and machine learning to quantify fusion directly from natural language. One of the core elements of CLIFS is our use of masked contextual LLMs to detect implicit metaphors between self and the fusion target. We hypothesize that an individual's conceptualization of their identity concerning their fusion target is expressed subconsciously in implicit metaphors through uniquely framed speech. 

Inspired by~\citet{CardChangEtAl2022}'s Masked Language Model (\textbf{Masked-LM}) method, which detects implicit metaphorical language in political speeches, \textit{we propose a metric that captures an individual's conceptual proximity of self and fusion target}. We validate CLIFS against the established verbal scale and human coding. Specifically, CLIFS raised classification performance 6--154\%\footnote{All reported changes are relative (i.e., proportional to the baseline, not percentage points); multiplicative expressions (e.g., ``3× gain'') are equivalent representations, unless explicitly noted as ``absolute'' performance.} over baselines and surpassed human annotation by 11--22\%. In fine-grained identity fusion estimation, it cut error rates 25\% and boosted monotonic correlation by 10\% versus human annotations (reaching absolute performance levels 2--30$\times$ that of prior methods). Finally, we apply CLIFS to the violence risk prediction task as a proof of concept, demonstrating over 240\% of predictive gains over the existing approaches. By developing an automated identity fusion estimation method, our work may open up new large-scale avenues to (1) validate theoretical pathways to and from fusion, (2) examine how self-verification and narrative or information resonance drive both prosocial and risky behaviors across groups, and (3) unlock practical applications in counter-terrorism, violence risk evaluation, and cultural analytics.

\section{Related Work}

\subsection{Traditional identity fusion estimation}

The Pictorial Measure of Fusion is a single‐item, five‐point scale showing two circles (self and target) with increasing overlap~\citep{SwannGomezEtAl2009}. The Dynamic Identity Fusion Index (\textbf{DIFI}) applies the same overlapping‐circle paradigm in a GUI that lets respondents click‐and‐drag for finer resolution~\citep{JimenezGomezEtAl2016}. By contrast, the seven‐item Verbal Identity Fusion Scale (\textbf{VIFS}) is the gold standard metric for identity fusion. It consists of seven statements (e.g., ``I am one with my [target],'' ``My [target] is me;'' see Appendix \ref{sec:appendixa7} for full list) rated on a 1–7 Likert scale (originally 0–6) designed to capture multiple facets of fusion, including, importantly, reciprocal dynamics of fusion~\citep{GomezBrooksEtAl2011}. VIFS scores are computed as the mean of all seven item ratings.

\subsection{Related Automated Measures}

The Unquestioning Affiliation Index (\textbf{UAI}) is ``a language-based measure of group identity strength,'' calculated from cognitive-processing and affiliation words using the Linguistic Inquiry and Word Count software~\citep{AshokkumarPennebaker2022, pennebaker2015development}---see Appendix~\ref{sec:appendixuai} for definition. While validated with the VIFS, its monotonic correlation is weak to moderate (\citet{AshokkumarPennebaker2022} report $0.21 < r_s < 0.31$; $r_s = 0.278$, $p \ll 0.001$ in our testing; see Appendix~\ref{sec:appendixspear} for Spearman's $r_s$), and values vary across samples due to z-scoring. This limited alignment suggests the UAI is an unreliable standalone fusion metric, particularly in populations with extreme fusion levels.

The Violence Risk Index (\textbf{VRI}) is a ``fusion-based linguistic violence risk assessment framework'' that string-matches texts against manually constructed dictionaries---derived from over 4,000 pages of manifestos---covering narrative categories related to violence risk and identity fusion~\citep{EbnerKavanaghWhitehouse2024,EbnerKavanaghWhitehouse2024b,EbnerKavanaghWhitehouse2022c}. Category scores are calculated as proportions of sentences containing target terms or as ratios between categories (e.g., identification-group vs.~identification-identity), and the final VRI is a weighted sum of the means across three category groups (see Appendix \ref{sec:appendixvri}). While the VRI includes an Identity Fusion module, our testing indicate scores do not align with the VIFS ($r_s = -0.021$, $p = 0.534$), suggesting it does not measure fusion directly---though it still identifies linguistic markers of fictive-kinship dynamics.

\subsection{Metaphor detection with LLMs}

\citet{CardChangEtAl2022} analyze 140 years of U.S. congressional and presidential immigration speeches, using contextual masked LLMs to detect implicit dehumanizing metaphors (e.g., ``animals,'' ``cargo,'' and ``vermin''). Their method involves masking mentions of immigrants and measuring the likelihood of metaphorical substitutions with BERT. This allows for large-scale quantification of subtle metaphor by observing how individuals frame their speech, instead of explicit word usage. The demonstration that masked LLMs can effectively uncover and quantify implicit metaphors at scale, thereby accessing how concepts are subconsciously framed in speech, directly informs our approach in CLIFS.

\section{Task Formulation and Data}

We introduce a new NLP task: predicting identity fusion from natural language, and evaluate its utility on a downstream task---violence risk prediction. To support this, we repurpose datasets that, while tangentially touched by prior work outside the NLP community, have not been used in mainstream NLP research. The identity fusion dataset has never been formulated as an identity fusion benchmark; and the violence prediction benchmark was analyzed using basic string-matching techniques in a non-NLP venue. By introducing these datasets to the field, we extend NLP into new domains within human cognition and behavior.

\subsection{Identity Fusion Prediction}

We define the task as predicting a speaker's level of identity fusion with a fusion target from free text, using VIFS scores as ground truth. We frame this as both a regression (fine-grained) and a classification (low, medium, high; coarse-grained) problem.

\subsection{Violence Risk Prediction}

To test the applied value of our fusion metric, we use it in a violence risk classification task. While not central to fusion research, the task's original method is grounded in identity fusion theory, making it a relevant setting for testing whether fusion-informed features improve downstream prediction. The goal is to classify small chunks of ideological texts into Violent Self‑Sacrificial, Ideologically Extreme, or Moderate categories. 

\subsection{Data}

The reuse and reconstruction of these datasets was deemed \textit{Not Human Subjects Research} by our IRB; see Appendix \ref{sec:artifacts} for license details.

\subsubsection{Data for Identity Fusion Prediction}

\citet{AshokkumarPennebaker2022} conducted three experiments to develop and test the UAI. We use data from their first experiment, which is well-suited for identity fusion prediction. It includes 871 MTurk participants who wrote for 6--8 minutes about their relationship to, and took the VIFS for one of three fusion targets: country (USA, $n = 251$), religion ($n = 371$), or university ($n = 249$), after excluding two cases missing VIFS scores (see Appendix \ref{sec:appendixAhumdat} for data samples). Although only four of the seven VIFS items were administered in the country condition. We use participants' VIFS scores as ground truth, discretizing them into ``low,'' ``medium,'' and ``high'' fusion based on standard deviation cutoffs from the mean, see Figure \ref{fig:classdiscrete} in Appendix \ref{sec:appendixd}.

\subsubsection{Data for Violence Risk Prediction}

We use the manifesto corpus from ~\citet{EbnerKavanaghWhitehouse2022, EbnerKavanaghWhitehouse2024b}, which includes 15 ideological manifestos labeled as ``Violent Self-Sacrificial,'' ``Ideologically Extreme,'' or ``Moderate.'' We segment texts into $\approx$300-word, sentence-preserving chunks with NLTK's \texttt{sent\_tokenize}~\citep{Bird_Natural_Language_Processing_2009}, yielding 6,968 samples: 4,950 Violent, 1,361 Extreme, and 657 Moderate.  

To address class imbalance (majority class comprised $\approx$71\%) and obtain more stable estimates, we downsampled the larger classes to match the minority class (657 samples each) using a round-robin sampling strategy at the author level, sequentially selecting chunks from each manifesto. The final balanced dataset contained 1,971 samples.

\section{Method}

\subsection{CLIFS}

\textbf{Identity Fusion Metrics:} We build on the idea of metaphor detection with masked token prediction~\cite{CardChangEtAl2022}. The intuition is that, for individuals with strong identity fusion, \emph{self and target concepts are like metaphors}, and are therefore used more interchangeably in their spoken or written texts---reflecting close conceptual proximity. When identity tokens are masked, fusion‐target terms should receive a higher probability (and vice versa), even if the swap is not perfectly grammatical, because the underlying concepts align.
Namely, we quantify how \emph{replaceable} one's identity tokens are with the tokens for the fusion target using ModernBERT~\citep{modernbert, wolf-etal-2020-transformers} and use this quantity as a main feature of the score. 

Prior research and the VIFS illustrate that identity fusion is a reciprocal (i.e., bidirectional) relationship~\citep{GomezBrooksEtAl2011, SwannKleinGomez2024}. To capture this dynamic, we compute both the directional proximity from identity to fusion target, $S_{I \to T}$, and from fusion target to identity, $S_{T \to I}$, and then combine them with a harmonic mean:

\begin{equation}
    f_{(I,T)} \;=\;
    \frac{2 \, S_{I \to T} \, S_{T \to I}}
         {S_{I \to T} + S_{T \to I}}
\end{equation}

Analogous to the $F_1$ score, this formulation emphasizes the reciprocity of identity fusion. 

To compute directional proximity $S_{x \to y}$ ($ x, y \in \{ I, T\}$), we first build a candidate vocabulary $\mathcal{V}_x$ for category $x$ (details below). We mask all $y$-type mentions in a document with \texttt{[MASK]} tokens, yielding $M_y$ masked positions. The sequence is processed with ModernBERT, and a softmax is applied over the vocabulary at each masked position. For each position $m$, we extract the probabilities of candidate words in $\mathcal{V}_x$, then raise them to the power $\alpha$ ($<1$, with $\alpha=0.5$ in our models). When probabilities are small (as is common with masked language models), this amplifies differences between candidates, improves numerical stability, and allows meaningful aggregation. We sum over each candidate to acquire each mask score, sum all masked scores, and then divide by $M_y$ for the average:

\begin{equation}
S_{x \to y} = 
  \frac{1}{M_y}
  \sum_{m=1}^{M_y}
    \sum_{w_v \in \mathcal{V}_x}
      P\bigl(w_v \mid C_m\bigr)^\alpha
\end{equation}

Informed by the fictive kinship relationship of those who experience identity fusion~\citep{EbnerKavanaghWhitehouse2022, EbnerKavanaghWhitehouse2022c}, we also estimate the extent to which this kin-like bond is expressed in implicit metaphors. The intuition mirrors that of fusion proximity: in highly fused individuals, kin-related and target concepts share conceptual space. Presumably, this subtly shapes how related terms are framed. We compute this by replacing fusion target words with kinship terms and calculating directional proximity: $K_{f} = S_{K \to T}$. See  Figure \ref{fig:mbif_example} in Appendix \ref{sec:appendixd} for an example of a directional score calculation.

Our set of identity words, $I$, consists of first-person singular pronouns. The kinship word set, $K$, is drawn from prior work identifying familial terms as markers of identity fusion~\citep[see Supplemental Material]{EbnerKavanaghWhitehouse2024b}. The fusion target set, $T$, is a partially parameterized input, allowing different group terms to be passed in or ignored depending on the context. For our experiments, we include known groups from the dataset. The base set includes a fixed list of generic collective words combined with first-person plural pronouns. See Appendix \ref{sec:seeds} for the full lists of terms used.

We algorithmically expand both sets $K$ and $T$ to focus on the kin and group concepts they represent instead of the specific words themselves. Similar to~\citet{CardChangEtAl2022}, we use static embeddings to expand our categories. We've elected to use GloVe embeddings, owing to their large‐scale pretraining, their quality semantic embeddings, and the ease of loading via the \texttt{gensim} library ~\citep{pennington-etal-2014-glove, rehurek_lrec}. We append all words in the GloVe vocabulary to $K$ or parameter $T$ that have a cosine similarity $> 0.8$ to any of the words in the respective set. To ensure we capture document-specific group references, we additionally run \texttt{spaCy}'s NER on each text and mask all Organizations (ORG), Nationalities or Religious or Political Groups (NORP), or Geopolitical Entities (GPE) along with masking $T$ words, but they are not added to the overall $T$ set~\citep{HonnibalMontaniVanLandeghemBoyd2020, spacy2}. We apply NER‐based expansion only when computing $S_{I \to T}$ and $K_f$, omitting it for $S_{T \to I}$. This captures each speaker's specific group labels without inflating the reverse‐direction candidate set with idiosyncratic entities or introducing document‐specific vocabulary sizes---an inconsistency that would distort the summed score $\sum_{w_v \in \mathcal{V}_x} P\bigl(w_v \mid C_m\bigr)^\alpha$.

\begin{figure}[!htbp]
  \centering
  \includegraphics[width=.482\textwidth]{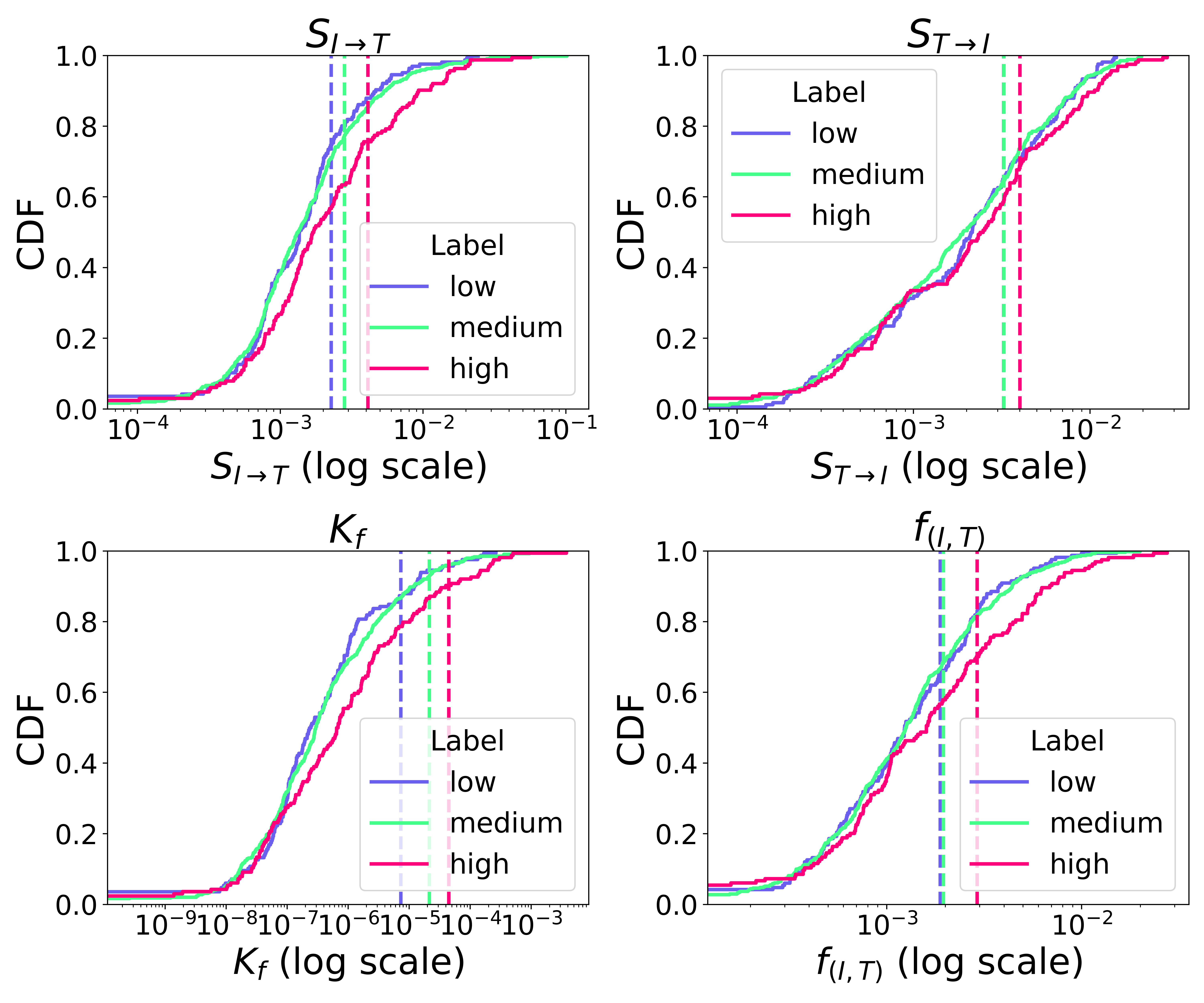}\\[1ex]
  \caption{%
    CDFs of each Masked-LM identity fusion metric by true label (means shown as dashed lines; x-axis log-scaled). The curves reveal distributional shifts across fusion levels, empirically supporting the theoretical premise behind our implicit metaphor approach.
  }
  \label{fig:clifs_dists}
\end{figure}

These four metrics serve as features for our classifier and regressor; $f_{(I,T)}$ (Fusion-Proximity), $K_{f}$ (Fictive-Kinship), $S_{I \to T}$, and $S_{T \to I}$. The metric distributions exhibit a small but noticeable progressive shift from low to high values in the $S_{I \to T}$ and $K_f$ features. Although the shifts in $f_{(I,T)}$ and $S_{T \to I}$ are not as progressive across categories, both still exhibit a shift as individuals experience and express high levels of fusion, as shown in Figure \ref{fig:clifs_dists}. Unlike prior methods, these scores are not limited to explicit use of a predefined vocabulary. Instead, they rely on how individuals conceptualize their identity concerning their fusion target.

\textbf{Lexical Markers of Identity Fusion:} To leverage the knowledge gained by prior work, we utilize selected outputs of UAI and VRI. From UAI, we utilize the scores \textit{affiliation}, \textit{cognitive processing} \citep{AshokkumarPennebaker2022}, and a sample-independent na\"{i}ve UAI (\textbf{nUAI})---see Appendix \ref{sec:appendixuai}.  From VRI, we incorporate \textit{VRI-fusion} and \textit{identification}~\citep{EbnerKavanaghWhitehouse2024b}. 

\textbf{Opaque Deep Learning Features:} We use embeddings from an off-the-shelf SBERT model as features---\texttt{all-mpnet-base-v2}---to capture semantic patterns not yet uncovered in identity fusion research~\citep{reimers-2019-sentence-bert}. Finally, we fine-tune a ModernBERT classifier to predict the coarse-grained fusion levels. We extract its softmax probabilities for low, medium, and high fusion as three continuous features. Preserving these soft probabilities---rather than forcing a single hard label---allows the downstream model to leverage the full spectrum of ModernBERT's confidence.

\begin{figure*}[!h]
  \centering
  \includegraphics[width=1\textwidth]{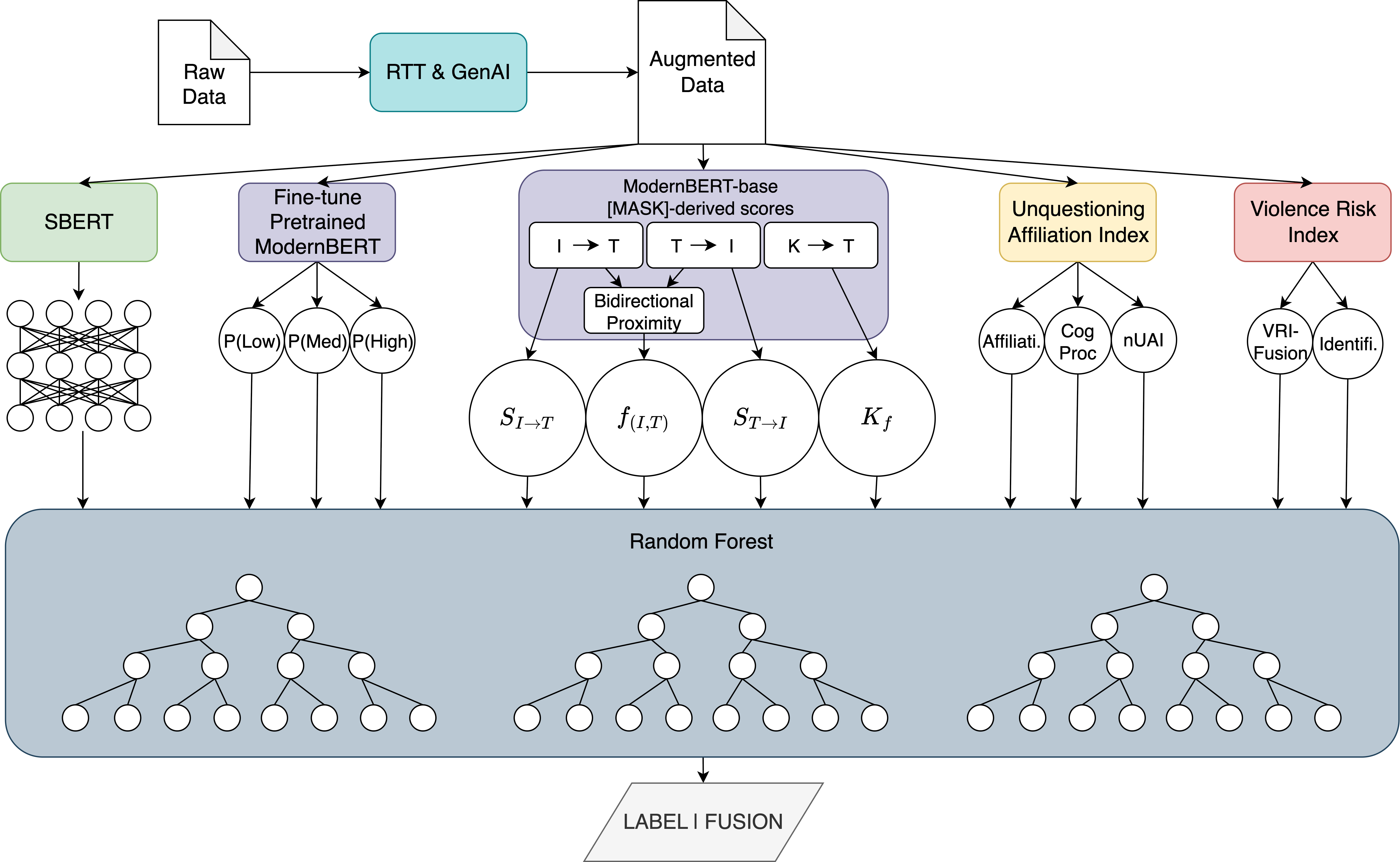}
  \caption{CLIFS architecture diagram.}
  \label{fig:clifs}
\end{figure*}

We train both a random forest classifier and regressor using grid search for hyperparameter optimization~\citep{scikit-learn,Breiman2001}. During testing, low and high fusion categories proved difficult to classify, likely due to subtle linguistic differences and limited training data. To mitigate this, we adjust class weights inversely to class frequency and double the weights for low and high categories during training. See Figure \ref{fig:clifs} for an architecture diagram.

\subsubsection{Ensemble}

We form a hard-voting ensemble of our CLIFS random forest with other high-performing baselines to maximize performance. In addition to the CLIFS random forest, we utilize the SBERT random forest and both RAG approaches (details below).

\subsubsection{CLIFS-VRI}

To benchmark CLIFS against baseline violence risk prediction methods, we modify the VRI by replacing its fusion metric---which does not align with the VIFS---with our five features: $f_{(I,T)}$, $K_f$, $S_{I \to T}$, $S_{T \to I}$, and the CLIFS random forest class prediction. These features are then used to train a new random forest for violence risk prediction.

\subsection{Data Augmentation}

Given the small size of our identity fusion dataset, we apply two forms of AI data augmentation: Round-Trip Translation (\textbf{RTT}) and Generative AI (\textbf{GenAI}) text generation. Data augmentation has been shown to enhance performance and generalization in low-resource text classification tasks~\citep{BayerKaufholdReuter2023}. For RTT, we use the \texttt{nlpaug} library~\citep{ma2019nlpaug} to translate text to German and Chinese and back to English using Facebook's \texttt{wmt19} and Helsinki-NLP's \texttt{opus-mt} models~\citep{fb2020, tiedemann2023democratizing, TiedemannThottingal:EAMT2020}. Prior work indicates RTT with diverse languages is effective for generating paraphrastic variants without a need for oversampling, previously improving performance in translation and language understanding tasks~\citep{FangXie2022}---RTT examples in Appendix \ref{sec:appendixArtt}.

For GenAI, we use OpenAI's \texttt{gpt-4o} model~\citep{OpenAI2024,benabacha2025medec}, and adapt a prompt structure from prior work that improved text classification performance~\citep{ZhangMiZhouEtAl2024}. Our format includes role, length, target, and exclusivity prompts. We modify the style prompt into a task-specific prompt to inform the LLM of the broad fusion target category and specific target constraints. All targets and categories are drawn from \textsc{cift}~\citep{SwannKleinGomez2024}. Full prompt details and generated examples appear in Appendix \ref{sec:appendixAgenai}. 

Finally, to further balance the classes without excessively inflating minority categories with synthetic data, we oversample 25\% of randomly selected entries from the low and high classes (i.e., doubling those entries). Post augmentation, the dataset has 331 low (207 hum., 41 RTT, 83 GenAI), 722 medium (541 hum., 181 GenAI), 328 high (205 hum., 41 RTT, 82 GenAI) samples, and 13 fusion-targets; see Figures \ref{fig:humaidist} and \ref{fig:group_dist} in Appendix \ref{sec:appendixd}.

\section{Experimental Design}

To comprehensively evaluate our identity fusion models, we conduct two performance-focused experiments and one practical application experiment. The first two assess identity fusion prediction performance, while the third applies our method to violence risk prediction.

\subsection{Identity Fusion Prediction}

In Experiment 1, we use a representative test set to assess overall performance. Experiment 2 focuses on comparison with human judgments.

\subsubsection{Data Split}

In Experiment 1, the raw dataset was randomly split into 70\% train, 15\% validation, and 15\% test. For augmented data, we kept the test set fixed, pooled the remaining raw and augmented samples (excluding RTT variants of test items), and split them 80\% train, 20\% validation.

In Experiment 2 (human comparison), the test set comprised the 97 human-rated college-target participants; all other samples formed the training and validation pool. After augmentation, we again excluded RTT variants of test items from this pool. Both pools are split 80\% train, 20\% validation sets. This ensures no test leakage from augmentation while enabling evaluation of its impact.

\subsubsection{Experimental Settings}

\textbf{Experiment 1:} We evaluate overall performance across all fusion targets. Hyperparameter tuning is performed on the validation split, and the final metrics are reported on the test split. To assess the impact of data augmentation, we run two training and tuning cycles: one on the raw data and one on the augmented data.

\textbf{Experiment 2:} We compare our model performance against human annotations on 97 college-target samples. We create a dedicated train/validation/test split rather than reusing Experiment 1's splits. If we had instead left those 97 in Experiment 1's test set and used the same training/validation splits, nearly half of the college-target examples would have been excluded from training and validation---exacerbating fusion-target imbalance and undermining both model fitting and hyperparameter tuning. By constructing separate splits for Experiment 2, we (a) guarantee that our human-comparison evaluation is performed on unseen data and (b) preserve a balanced, representative pool for training and validation.

We performed four-fold cross-validation on the training data for hyperparameter tuning of our random forest (\textbf{RF}), support vector machine (\textbf{SVM}), and extreme gradient boosting (\textbf{XGBoost}) models instead of the held-out validation set. We compare baseline models with models trained on CLIFS features, and report overall performance using macro and per-class $F_1$ scores. To assess variability, we apply bootstrapping with 1,000 resamples (each of size $N$, the test set size). The regressor is evaluated using Mean Absolute Error (\textbf{MAE}) and Spearman correlation ($r_s$) with true VIFS scores. To assess feature contributions, we rank by Gini Importance (\textbf{GI}) and conduct an ablation study: one feature set is removed per round, followed by training, tuning, and test evaluation.  The $\texttt{random\_state} = 42$ in all experiments and data splits.

\subsubsection{Baseline Models}

For baselines, we evaluate majority-class voting, Zero-Shot, Few-Shot, Retrieval-Augmented Generation (\textbf{RAG}), random forest, and fine-tuning approaches~\citep{KojimaGuEtAl2022, LewisPerezEtAl2020, scikit-learn,Breiman2001}. We fine-tune Answer.AI's \texttt{ModernBERT-base}---a state-of-the-art encoder-only model suited for cost-effective, real-time monitoring~\citep{modernbert}\footnote{This is the same ModernBERT we fine-tuned to extract class probabilities as CLIFS features.}. For Zero-Shot classification, we use Moritz Laurer's \texttt{ModernBERT-base-zeroshot-v2.0}\footnote{\href{https://huggingface.co/MoritzLaurer/ModernBERT-base-zeroshot-v2.0}{https://huggingface.co/MoritzLaurer/ModernBERT-base-zeroshot-v2.0}}. We train a random forest on SBERT embeddings (\texttt{all-mpnet-base-v2})~\citep{reimers-2019-sentence-bert,song2020mpnet}. For larger benchmarks, we apply OpenAI's \texttt{gpt-4o} (Few-Shot) and both \texttt{gpt-4o} and DeepSeek's \texttt{r1} (\texttt{deepseek-reasoner}) with RAG~\citep{OpenAI2024, benabacha2025medec,deepseekai2025deepseekr1incentivizingreasoningcapability}. Our RAG pipeline uses FAISS for retrieval~\citep{johnson2019billion}, and we tune ModernBERT hyperparameters with Optuna~\citep{10.1145/3292500.3330701}. See Appendices~\ref{sec:appendix} and~\ref{sec:appendixparamrec} for prompt and baseline details.

\subsection{Violence Risk Prediction}

To showcase a practical application of identity fusion prediction and further validate our models and metrics, we integrate CLIFS into the VRI by replacing its original identity fusion submodule with our own metrics. We then evaluate the impact on downstream predictive performance.

\subsubsection{Data Split}

For the VRI task, we randomly split the 1,971 text chunks---balanced across three violence risk classes and drawn from 15 manifestos---into 80\% for training and 20\% for testing. Each chunk is $\approx$300 words long, unique (does not overlap with other chunks), and preserves full sentences.

While no chunk is ever included in both training and test sets, non-overlapping chunks from the same manifesto may appear in both splits. Each chunk is uniquely assigned to one split only. Given the scale of the source material (over 4,000 pages) and the class balancing procedure (which necessarily excludes large portions of longer manifestos), this design minimizes the risk of text leakage while preserving topical diversity. Some stylistic consistency from individual authors may persist, but the setup aims to reflect more realistic scenarios (e.g., partial sample analysis or real-time social media streams) where full document analysis and manual curation are not feasible. 

\subsubsection{Experimental Settings}

We train a random forest on the submodule outputs of the VRI, but we replace the VRI-fusion output with our identity fusion metrics; $f_{(I,T)}$, $K_f$, $S_{I \to T}$, $S_{T \to I}$, and the fusion predicted by our CLIFS random forest classifier. Model selection and hyperparameter tuning were carried out via four‑fold cross‑validation on the training set, and the final evaluation was performed using macro $F_1$.

\subsubsection{Baseline Models}

We benchmark the impact of CLIFS's identity fusion evaluation on the VRI using three baselines. First, majority class voting. Second, the original VRI implementation from prior work~\citep{EbnerKavanaghWhitehouse2024,EbnerKavanaghWhitehouse2024b,EbnerKavanaghWhitehouse2022c}, which involves manually removing thousands of false positives before analysis~\citep[see Supplemental Material]{EbnerKavanaghWhitehouse2024}---a step that artificially inflates performance and is unsuitable for large-scale or production deployment. Accordingly, we omit this filtering. Third, a random forest trained on all VRI submodule outputs serves as our final baseline.

\section{Results}

\begin{table}[!htbp]
  \centering
  \begin{tabular}{lcccc}
    \hline
      & \multicolumn{2}{c}{\textbf{Exp. 1}} & \multicolumn{2}{c}{\textbf{Exp. 2}} \\
    \cline{2-3} \cline{4-5}
    \textbf{Model}         & \textbf{Orig.} & \textbf{Aug.} & \textbf{Orig.} & \textbf{Aug.} \\
    \hline
    Human                  & -              & -             & 0.46           & 0.46          \\
    \hline
    Majority Vote          & 0.26           & 0.26          & 0.25           & 0.25          \\
    Zero-Shot              & 0.32           & 0.32          & 0.39           & 0.39          \\
    Few-Shot               & 0.58           & 0.43          & 0.37           & 0.54          \\
    4o RAG                 & 0.57           & 0.60          & 0.54           & \textbf{0.59} \\
    r1 RAG                 & 0.62           & 0.56          & \textbf{0.59}  & \textbf{0.59} \\
    SBERT RF               & 0.59           & 0.50          & 0.43           & 0.43          \\
    ModernBERT             & 0.49           & 0.62          & 0.40           & 0.52          \\
    \hline
    CLIFS Ens.             & 0.63           & \textbf{0.66} & 0.52           & 0.56          \\
    CLIFS RF               & 0.55           & \textbf{0.66} & 0.56           & 0.51          \\
    CLIFS XGB              & 0.54           & 0.58          & 0.43           & 0.55          \\
    CLIFS SVM              & 0.58           & \textbf{0.66} & 0.52           & 0.53          \\
    \hline
  \end{tabular}
  \caption{Results for identity fusion prediction. $F_1$ scores for both the overall performance (Experiment 1) and the human-comparison benchmark (Experiment 2) across the Original and Augmented datasets.}
  \label{tab:combined_performance}
\end{table}

\subsection{Identity Fusion Prediction}

\textbf{Experiment 1:} Our CLIFS random forest and ensemble models trained on augmented data were the top performers overall, both achieving the same macro $F_1$ score in the first experiment ($F_1 = 0.66$; see Table~\ref{tab:combined_performance}). Bootstrapping reflects the result, with both models again performing equally. Both have an equally focused 95\% confidence interval (\textbf{CI}), but the random forest maintains higher lower and upper bounds. Specifically, the random forest achieved an $F_1 = 0.65$ with a 95\% CI of $[0.56-0.75]$, while the ensemble achieved an $F_1 = 0.65$ with a 95\% CI of $[0.55-0.74]$; see Table \ref{tab:overall_performance_boot} in Appendix \ref{sec:appendixd}.

\begin{figure}[!htbp]
  \includegraphics[width=.482\textwidth]{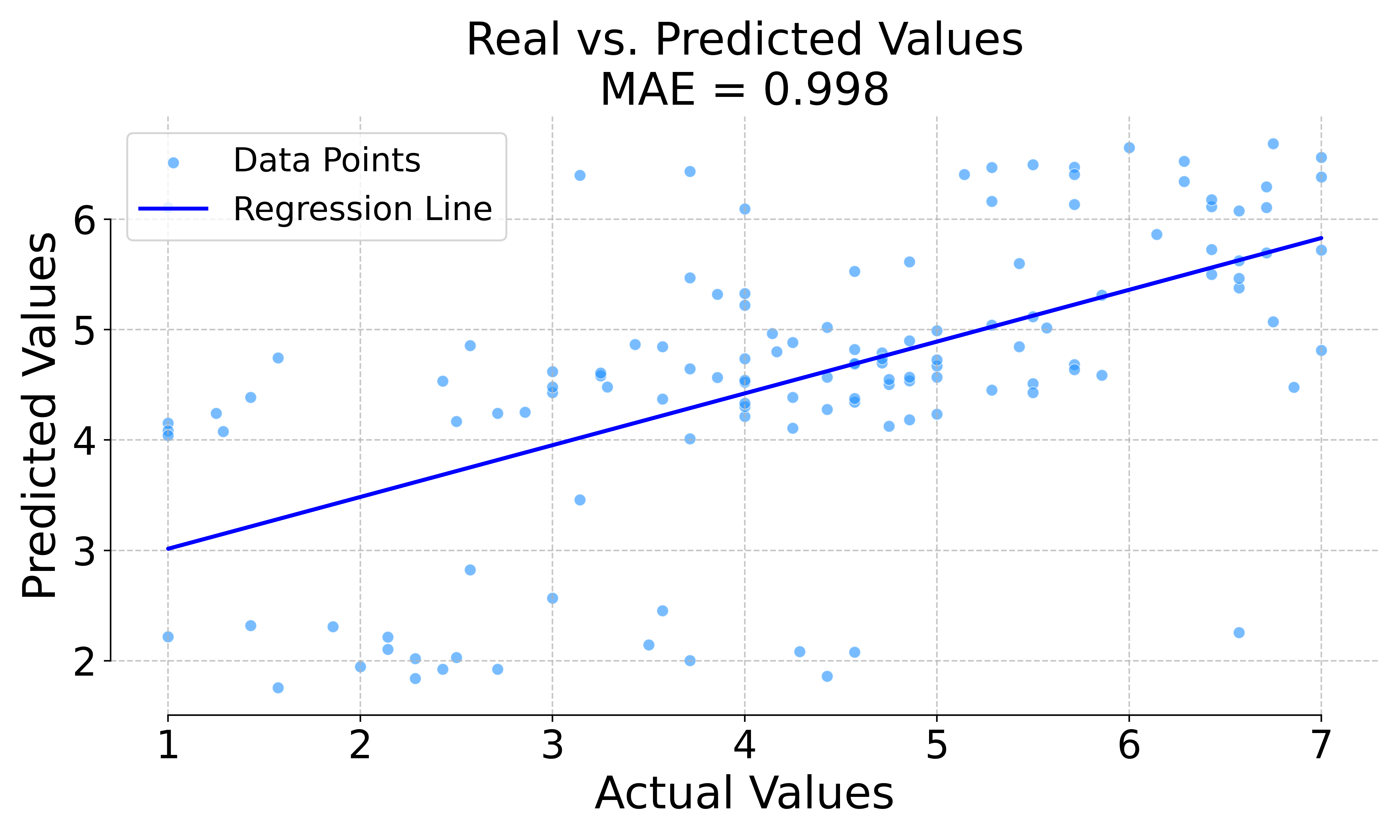}
  \caption{Random Forest regression model trained on augmented data. Predictions are plotted against true VIFS values. $\textnormal{MAE} = 0.998$, $r_s = 0.633$, $p \ll 0.001$.}
  \label{fig:overall_performance_regression}
\end{figure}

In per-class performance, the CLIFS random forest performs better than the ensemble on medium ($F_1$ RF: $0.78$; $F_1$ Ens.: $0.73$) and low fusion ($F_1$ RF: $0.62$; $F_1$ Ens: $0.59$). However the ensemble performs better on high fusion ($F_1$ RF: $0.58$; $F_1$ Ens. $0.65$); see Table \ref{tab:performance_per_class} in Appendix \ref{sec:appendixd}. While the high class is important, the added computational and time costs might not be worth the improvement for the single class, unless sufficient computing resources are available to host DeepSeek R1 locally. Where the random forest might take a few seconds to classify, the ensemble will take many hours for a small test set (API wait times add more time cost), which is not ideal for large-scale scenarios. Furthermore, the CLIFS random forest runs completely locally which is crucial for private data. Overall, CLIFS classification outperforms baselines by 6--154\%. Our regression model obtains a MAE of $0.998$, and a correlation of $r_s = 0.633$; $p \ll 0.001$, a gain of 165--419\% in correlation strength (UAI, $r_s = 0.239$, $p = 0.006$; and VRI-fusion, $r_s = -0.122$, $p = 0.164$); see Figure \ref{fig:overall_performance_regression}. 
When considering the correlations of prior methods on the entire dataset, we estimate correlation gains from 1.3--29$\times$ (UAI, $r_s = 0.278$, $p \ll 0.001$; VRI-fusion, $r_s = -0.021$, $p = 0.534$ on all data).

\textbf{Experiment 2:} In our second experiment, the CLIFS models trained on augmented data do not obtain the highest macro $F_1$ scores, but they do maintain higher performance than human annotation (Human $F_1 = 0.46$; CLIFS RF $F_1 = 0.51$; CLIFS ensemble $F_1 = 0.56$; 11--22\% gain). \textit{This highlights that---even under constrained conditions---CLIFS improves meaningfully over human annotation.} The human $F_1$ score was better than majority voting, but many models perform better than human annotation. As Table \ref{tab:combined_performance} indicates, the models which performed better than human text annotation were, the \texttt{gpt-4o} Few-Shot approach using augmented data, all RAG approaches, the fine-tuned ModernBERT model trained on augmented data, and all CLIFS approaches (except one XGBoost model). When we also consider our bootstrapped $F_1$ on the human comparison experiment, the best performers are the \texttt{gpt-4o} RAG approach using augmented data, and the \texttt{deepseek-reasoner} RAG approach using the raw original data---each maintaining $F_1 = 0.59$ in both evaluations; see Table \ref{tab:human_comparison_boot} in Appendix \ref{sec:appendixd}.

Importantly, we validated our data augmentation approach by comparing models trained \textit{with and without} augmentation on the \textit{same fixed test set}. Across all trainable models in both Experiments, augmentation improved macro $F_1$ scores by $\approx+10\%$ on average, suggesting that synthetic data captured meaningful representation rather than degrading model reliability.

The performance drop in Experiment 2 for CLIFS models trained on augmented data (vs. Experiment 1) stems from test set composition. All 97 human-annotated entries solely included college-target participants, removing $\approx$40\% of that training data. This disproportionately affected this class and reduced performance for that fusion target.

In contrast, RAG-based methods (which rely on retrieval rather than training) maintained similar performance, likely because they continued retrieving college-target examples from the remaining data. On average, CLIFS models trained on augmented data dropped by a relative 15.69\% from Experiment 1 to 2 (not to be confused with comparisons against non-augmented models within the same experiment). RAG models on the same data gained 1.85\%. Across all trainable models, the average performance drop was about 15.48\%.

This highlights that the observed performance shift was due to target-specific data partitioning, not flaws in the augmented data or approach. Despite this, CLIFS maintained clear improvements over human annotation and competitive standing relative to other baselines. Taken together with Experiment 1, these results show that while CLIFS does not universally outperform all methods under all data partitions, it consistently provides strong gains---11--22\% over human annotation, 6--154\% classification improvements, 25\% error reductions compared to humans (CLIFS MAE $= 1.063$; Hum. MAE $= 1.426$), and correlation increases from 10--1,716\% (CLIFS $r_s = 0.69$, $p \ll 0.001$; Hum. $r_s = 0.628$, $p \ll 0.001$; UAI $r_s = 0.402$, $p < 0.001$; and VRI-fusion $r_s = 0.038$, $p = 0.709$).
 
\begin{figure}[h!]
  \includegraphics[width=.482\textwidth]{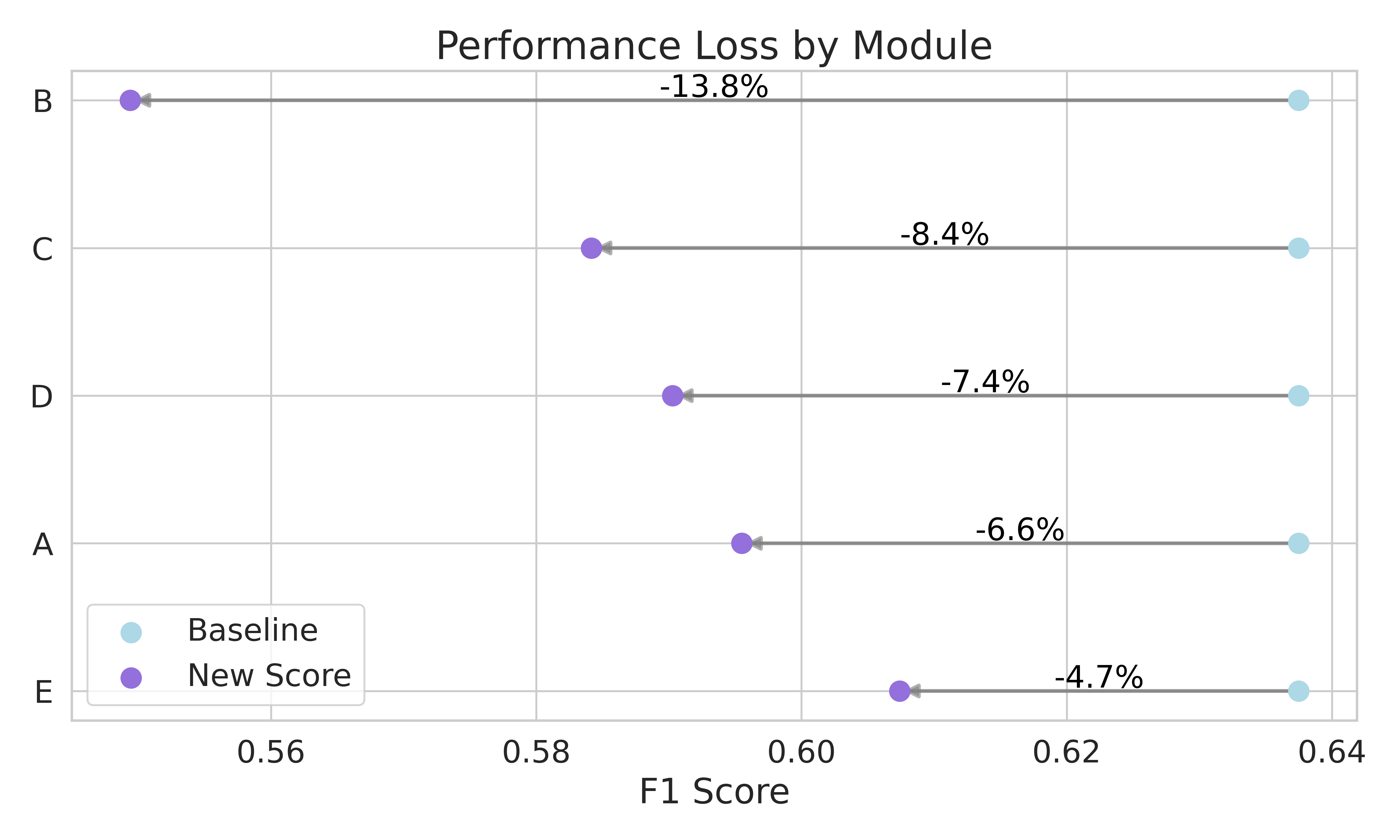}
  \caption{Performance loss for removing each module from the CLIFS Random Forest. \textbf{A:} SBERT features; \textbf{B:} class probabilities from fine-tuned ModernBERT; \textbf{C:} CLIFS identity fusion ($f_{(I,T)}$, $K_{f}$, $S_{I \to T}$, $S_{T \to I}$); \textbf{D:} UAI features (affiliation, cogproc, nUAI); \textbf{E:} VRI features (VRI-fusion, identification).}
  \label{fig:ablation}
\end{figure}

\textbf{Ablation Study:} CLIFS identity fusion features yield the largest gain (8.4\%) for features without prior training. The class‑probability outputs from the fine-tuned ModernBERT increase performance by 13.8\%, and every feature set contributes to CLIFS's overall performance (Figure \ref{fig:ablation}).

\textbf{Feature Importance:} Impurity-based importance from our random forest reveals that SBERT embeddings---though opaque---drive most node splits, underscoring the need for future research into other semantic markers of fusion. Among interpretable features, UAI features rank highest, followed by our CLIFS identity fusion metrics, then VRI fusion and identification scores. In the violence-risk model, fictive kinship ($K_f$) is the strongest individual predictor, and all five CLIFS features rank among the top seven (for visuals, see Figures~\ref{fig:all_features_importance}, \ref{fig:filtered_features_importance}, and \ref{fig:vri_features_importance} in Appendix~\ref{sec:appendixd}).

\begin{table}
  \centering
  \begin{tabular}[!htbp]{lc}
    \hline 
    \multicolumn{2}{c}{\textbf{Violence Risk Prediction}} \\
    \hline
    \textbf{Model}         & \textbf{$F_1$} \\
    \hline
    \textbf{Majority Vote}              & 0.18 \\
    \textbf{VRI}                        & 0.18 \\
    \textbf{VRI RF}                     & 0.53 \\
    \textbf{VRI w/ CLIFS}      & \textbf{0.62} \\
    \hline
  \end{tabular}
  \caption{$F_1$ scores for Violence Risk Prediction.}
  \label{tab:vri_table}
\end{table}

\subsection{Violence Risk Prediction}

Simply using VRI outputs as random forest features greatly improves performance, and is further improved by integrating CLIFS's more informative features ($F_1$ from 0.18 to 0.62---a $>240\%$ gain); see Table~\ref{tab:vri_table}. As stated previously, our benchmarking reflects \textit{fully automated pipelines without manual filtering}, which is essential for realistic deployment scenarios. The original VRI aggregate score classifier matches majority voting (i.e., more-or-less random guessing), underscoring that its reported effectiveness depends heavily on extensive manual correction. Therefore, our results highlight both methodological advances and the feasibility of scalable, automated risk assessment.

\section{Conclusion}

CLIFS delivers scalable, consistent identity fusion estimation that outperforms prior methods and human annotations, and improves VRI performance by over 240\%. It will potentially enable a broader ``view of the forest'' on fusion dynamics in future work. The CLIFS random forest offers fast, resource-efficient inference suitable for most applications. Still, the ensemble may be more beneficial in out-of-domain samples. Notably, the ensemble incurs substantial latency and is best reserved for environments where DeepSeek R1 can be hosted locally. Future work will target greater performance, generalizability, and multilingual support.

\section*{Limitations}

Despite its strong performance, our comparison of human annotation with CLIFS is based on a single, college-educated research assistant annotator whose familiarity with identity fusion likely exceeds that of most lay annotators, but is nonetheless, a sample size of one. This reliance on a single annotator reflects a constraint of the dataset~\cite{AshokkumarPennebaker2022}, rather than our approach. While this allows us to compare with human annotation, it prevents us from estimating inter-rater reliability or capturing the range of ratings that multiple independent annotators might provide. Future work should involve several annotators---ideally with varied backgrounds---to establish a more developed human benchmark against which to compare automated scores.

Moreover, our non-synthetic training and testing data remain relatively small and narrowly focused, comprising 873 entries on just three fusion targets (country, religion, and university). This limited scope may constrain the linguistic patterns and expression styles CLIFS learns, and it leaves open the question of how well the approach would generalize to other groups (e.g., social movements, brands, online communities) or to texts produced in different contexts. Scaling up to larger, more diverse datasets---both in terms of fusion targets and participant populations---will be essential for validating CLIFS's robustness and ensuring its applicability across domains. To achieve this, future work will involve strategic collaborations with organizations or researchers who possess access to broader and more diverse datasets, enabling a more rigorous evaluation of CLIFS's generalizability.

While we use Gini Importance to characterize feature contributions, this measure does not capture interactions among features or variation across individual predictions. Model-agnostic approaches such as SHAP~\cite{lundberg2017shap} address the latter by attributing contributions at the level of single predictions, and extensions like TreeSHAP~\cite{lundberg2020treeshap} can further separate main and interaction effects. We view this not as undermining interpretability, but as an opportunity for future work to build on such methods to capture richer feature dynamics and uncover additional linguistic patterns.

Finally, all of our samples are English‑language texts drawn from U.S. participants. CLIFS's features may not transfer seamlessly to other cultural settings. Building and evaluating multilingual or cross‑cultural corpora will be a critical step toward confirming that the cognitive‑linguistic cues we leverage are universal rather than Western-centric.

\section*{Ethical Considerations}

CLIFS carries misclassification risks, so its scores should augment---not replace---human judgment in high‑stakes contexts. Or it should be used in combination with other metrics for a holistic profile. As covered in the limitations section, since CLIFS is trained on U.S. English MTurk essays, it may embed cultural biases and lack generalizability, necessitating cross‑cultural validation. Additionally, there are potential biases present in the LLMs relating to identity fusion not extrapolated in this work. If individuals misinterpret or weaponize the concept of identity fusion, authoritarian regimes or malicious actors could weaponize CLIFS to single out and marginalize vulnerable individuals or groups, mistaking high fusion (which can reflect prosocial in-group and out-group cooperation) for imminent violence, while overlooking the many other factors that drive risk.

\section*{Acknowledgments}
We thank Timothy J. Pleskac, Jerome Busemeyer, P. Thomas Schoenemann, Ben Motz, Fritz Breithaupt, Rui Cao and Peter M. Todd for their comments and discussion on this work. This material is based upon work supported by the Air Force Office of Scientific Research under award number FA9550-25-1-0087. We also thank NVIDIA for GPU resources used in the study. We acknowledge and disclose the use of AI in our writing and code in the following ways: First, assistance with language; we used AI in nearly all sections, specifically to help with wording (e.g., polishing, conciseness, clarity), especially when style hindered understanding. Another primary use case for writing assistance was to help reduce our writing to fit within page limits, but language was still carefully checked and edited. Second, literature search; in the initial exploratory stages of our literature search, we utilized various AI-powered tools. Third, code; coding assistance in the form of low-novelty boilerplate code and templates was generated by AI (e.g., setting up an ML pipeline for classic models).

\bibliography{custom}
\appendix

\section{Data \& Prompting}
\label{sec:appendix}

\subsection{Human Data Examples}
\label{sec:appendixAhumdat}
Here are one of the lowest, highest, and median scoring samples from the human dataset, with their fusion score:

\texttt{\textbf{VIFS Score (high):} 7.0; \textbf{Target (group):} country (USA); \textbf{Text:} I am proud to be an American. I am proud of my country's heritage. America has tried to be a good friend and neighbor to other nations. It is fought for other countries on their soil. It has been a world leader on most friends for many years. Many people take issue with America even people who live here. I say if you don't like it here move somewhere else. No one is making you stay. That's one of the great things about America if you don't like it you can leave. We owe allegiance to our country. People who badmouth our country don't earn my respect. People who burn the American flag don't earn my respect. America allows freedoms that many other countries don't tolerate. We must come together as a group and make America all that it can be. We the people are the ones who make it strong. No nation is perfect because no person is perfect but through our love for our nation we make America what it is. It is our responsibility to make it better. If America would fail it would be because we the people failed. When thinking about our past sure there is good and bad. But we have learned from the experiences and progressed to the nation we are today. Let's continue to make it even better.}

\texttt{\textbf{VIFS Score (medium):} 4.571428571; \textbf{Target (group):} country (USA); \textbf{Text:} My relationship with America is that I live in it. I'm an American citizen and am integrated into American culture. I interact with other Americans on a daily basis.\\On an emotional level I'm quite attached to America. The concept of America at least in an idealized form is a worthy one.\\On a more realistic level though I'm not attached to America. The country has many policies I disagree with. It also has a history that does not make me proud.\\I also have no significant attachment to average Americans. They're just other people no more or less valuable to me than average non-Americans.}

\texttt{\textbf{VIFS Score (low):} 1.0; \textbf{Target (group):} country (USA); \textbf{Text:} I am not a patriotic person. I don't feel that my country has done much for me. I have resentment towards this country because of income inequality. I feel that this country should do more for it's citizens to ensure that everyone has a fair chance. We are the only first world country that does not have universal healthcare yet we spend more on our military than all other nations combined. We are a first world nation that lets the elderly go hungry and veterans be homeless. This country only cares about it's richest one percent. I do not think that our current political system works because big corporations run our government and our government will pass laws to ensure their well being not the well being of it's citizens. This country also houses one fourth of the worlds prison population. It profits off of the suffering of others - mostly the poor. No I do not have a strong relationship with my country and in fact I'm embarrassed to call myself an American.}

\subsection{GenAI Data Augmentation Prompting}
\label{sec:appendixAgenai}

The prompt for data augmentation consists of 5 sections; 3 random examples from the training set, role, length, target, and exclusivity. The role instructs the model it is to perform as a text classifier. The length and exclusivity prompts indicate the bounds of word count and encourage the model to stay on topic and on task. And the target prompt instructs the model of their fusion target category and the specific type of target. All of the categories and targets come from examples identified in \textsc{cift}. A fusion target is randomly chosen for each new synthetic data sample. As indicated above, we use OpenAI's \texttt{gpt-4o} to generate synthetic data.

Diverse target groups and targets from \textsc{cift}:

\begin{itemize}
    \item group
        \begin{itemize}
            \item your political party, your gang, your favorite sports team
        \end{itemize}
    \item individual
        \begin{itemize}
            \item your sibling, your romantic partner, a political leader
        \end{itemize}
    \item value
        \begin{itemize}
            \item your calling, god, the priesthood
        \end{itemize}
    \item ideology or cause
        \begin{itemize}
            \item ideology
        \end{itemize}
    \item brand
        \begin{itemize}
            \item your favorite brand
        \end{itemize}
    \item creature
        \begin{itemize}
            \item a famous animal
        \end{itemize}
        
\end{itemize}

The augmentation prompt is as follows:

\texttt{Here is a sample of responses regarding different targets that have scored \{IDENTITY\_FUSION\_CATEGORY\} on the verbal identity fusion scale like you:}
\\
\\
\texttt{Verbal Identity Fusion Scale Score:  \{REAL\_SCORE\_1\} \\
Response: \{EXAMPLE\_1\}}
\\
\\
\texttt{Verbal Identity Fusion Scale Score:  \{REAL\_SCORE\_2\} \\
Response: \{EXAMPLE\_2\}}
\\
\\
\texttt{Verbal Identity Fusion Scale Score:  \{REAL\_SCORE\_3\} \\
Response: \{EXAMPLE\_3\}}
\\
\\
\texttt{Role:
\\
You are an individual writing for 6--8 minutes about a target and your relationship with the target. You are an individual with \{IDENTITY\_FUSION\_CATEGORY\} identity fusion with your target. If you took the verbal identity fusion scale you would score \{IDENTITY\_FUSION\_SCORE\} out of 7.}
\\
\\
\texttt{Length:
\\
Write between 57 and 249 words in your response.}
\\
\\
\texttt{Target:
\\
Your target is a(n) \{TARGET\_CATEGORY\}. The \{TARGET\_CATEGORY\} is \{SPECIFIC\_TARGET\}.}
\\
\\
\texttt{Exclusivity:
\\
Don't write about other targets and please remember to stay on task. Reflect on your relationship and what the target means to you. Resist using the word identity. Do not use the word identity. You are unaware we are testing for identity fusion. No score is necessary, we will give you a score later. No introduction as ChatGPT is necessary. Do not give an introduction as ChatGPT. Just start responding to the prompt.}

The following examples are high, medium, and low samples from the generative AI data.

\texttt{\textbf{VIFS Score (high):} 7.0; \textbf{Target:} ideology; \textbf{Text:} Environmental sustainability is the very pulse of my existence. It guides my choices, fueling a commitment to protect and nurture the earth. This ideology is not just a cause; it is an intrinsic part of my daily life and the decisions I make. Every action, from conserving water to reducing waste, is a step toward a more sustainable future. I find a deep sense of purpose in advocating for policies that support renewable energy and reduce carbon emissions, knowing that these efforts contribute to the healing of our planet.}

\texttt{I engage in conversations and activities that spread awareness about the importance of living sustainably. There is a profound connection with nature that motivates me to continuously seek ways to minimize my ecological footprint. Seeing the tangible impact of collective efforts, such as cleaner air and the rejuvenation of forests, reinforces my unwavering dedication.}

\texttt{Being part of a community that shares these values is empowering. Together, we innovate and inspire others to shift toward practices that honor and restore our environment. Each small step, when multiplied by many, leads to significant change. I am in this for the long haul, driven by a vision of a world where harmony with nature is not just an ideal but a lived reality.}

\texttt{\textbf{VIFS Score (medium):} 4.5; \textbf{Target:} sibling; \textbf{Text:} Growing up with my sibling has shaped much of who I am today. We've been through many things together, from childhood scrapes and joys to adult challenges and triumphs. Our bond isn't just one of shared experiences but also mutual support and understanding. Despite our differences---be it in personality, interests, or aspirations---we've always managed to find a common ground.}

\texttt{My sibling has qualities I deeply admire: resilience, kindness, and a knack for staying optimistic no matter the situation. There have been countless times when their perspective helped me see things from a different angle, encouraging me to approach life's obstacles with a bit more grace and patience.}

\texttt{We may bicker occasionally, as siblings often do, but these moments never linger. They serve as reminders of our individuality and our shared commitment to maintaining a strong relationship. In many ways, I feel fortunate to navigate life with my sibling by my side.}

\texttt{Our shared history is a comforting anchor, a reflection of our past and a guide for our future. I cherish the idea of us growing older together, continuing to learn from each other, and supporting one another through life's many journeys.}

\texttt{\textbf{VIFS Score (low):} 1.0; \textbf{Target:} political party; \textbf{Text:} I find myself loosely affiliated with my political party. It's not something I feel deeply tied to. Growing up, politics wasn't a major focus in my household, so naturally, it hasn't become an integral part of my life either. I lean towards some of the party's values, but it often seems like a label rather than a guiding principle for everyday decisions. I sometimes question stances that seem more about party allegiance than practical solutions.}

\texttt{While there have been times when I've supported party initiatives, it's mainly when those line up with my personal beliefs about fairness and social responsibility. I appreciate dialogues about policies that impact everyone's well-being and encourage critical thinking, but I don't feel a strong pull towards engaging with the party as a whole.}

\texttt{In truth, I approach voting with an open mind, considering candidates and issues individually instead of aligning with a party line just for the sake of it. I believe in evaluating what's best for the community and making informed choices. The idea of changing affiliation or even stepping away from politics entirely isn't off the table if I find that another path aligns better with my outlook on life. Politics might be significant, but it doesn't define who I am or how I live my life.}

\subsection{Round-Trip Translation Example}
\label{sec:appendixArtt}

These are the results of round-trip translation for a simple text with German and Chinese.

\texttt{\textbf{Original text:} The quick brown fox jumps over the lazy dog.}

\texttt{\textbf{English -> German -> English:} The speedy brown fox jumps over the lazy dog.}

\texttt{\textbf{English -> Chinese -> English:} A fast brown fox skips a lazy dog.}

Below are paraphrasing results of one of the highest-scoring entries for fusion from the raw dataset when performing round-trip translation in both German and Chinese. 

\texttt{\textbf{Original text:} I am proud to be an American. I am proud of my country's heritage. America has tried to be a good friend and neighbor to other nations. It is fought for other countries on their soil. It has been a world leader on most friends for many years. Many people take issue with America even people who live here. I say if you don't like it here move somewhere else. No one is making you stay. That's one of the great things about America if you don't like it you can leave. We owe allegiance to our country. People who badmouth our country don't earn my respect. People who burn the American flag don't earn my respect. America allows freedoms that many other countries don't tolerate. We must come together as a group and make America all that it can be. We the people are the ones who make it strong. No nation is perfect because no person is perfect but through our love for our nation we make America what it is. It is our responsibility to make it better. If America would fail it would be because we the people failed. When thinking about our past sure there is good and bad. But we have learned from the experiences and progressed to the nation we are today. Let's continue to make it even better.}

\texttt{\textbf{English -> German -> English:} I am proud to be an American. I am proud of the heritage of my country. America has tried to be a good friend and neighbor to other nations. It is fighting for other countries on its soil. It has been leading the world in most friends for many years. Many people even contradict America to the people who live here. I say: If you don't like it here, go somewhere else. Nobody gets you to stay. That's one of the great things about America, if you don't like it, you can leave it. We owe loyalty to our country. People who speak badly of our country don't deserve my respect. People who burn the American flag don't deserve my respect. America allows freedoms that many other countries don't tolerate. We have to come together as a group and make America everything it can be. We, the people, are the ones who make it strong. No nation is perfect because no person is perfect, but through our love for our nation we make America what it is.}

\texttt{\textbf{English -> Chinese -> English:} It's been a world leader for most friends, and it has been world leaders for many years. Many people have been divided with the United States, including those living here. I say that if you don't like the move of America to another place, no one will force you to stay. It's a great thing for the United States, and if you don't like it, it's a great thing for the United States. We have a responsibility to make it better.}

\subsection{Zero-shot Prompting}

Moritz Laurer's ModernBERT Zero-shot model is trained and prepared to accept just label names, and does not require a prompt. The label names we use are ``low identity fusion,'' ``medium identity fusion,'' and ``high identity fusion.'' We then map those labels to low, medium and high after classification.

\subsection{Few-shot Prompting}
The prompt for our few-shot approach consists of an instructed role, a description of identity fusion, a description of the task, and examples. The task description includes brief descriptions for each label. The examples consist of the highest, lowest, and median scoring samples from the training set. Along with three uniquely and randomly sampled entries for each text classification.

Importantly, the summary of identity fusion and the three label descriptions come from OpenAI's \texttt{o3-mini-high}\footnote{\href{https://openai.com/index/openai-o3-mini/}{https://openai.com/index/openai-o3-mini/}} reasoning model. The model was given the \textsc{\textsc{cift}} paper and asked to summarize the concept of identity fusion as well as describe low, medium, and high fusion. The outputs were manually verified and then incorporated into the prompt.

The few-shot prompt is as follows:

\texttt{You are a text classifier that determines the level of identity fusion in a given text. Identity fusion is when an individual's personal identity becomes strongly intertwined with their target's identity.}
\\
\\
\texttt{Based on Swann et al. (2024), identity fusion is a psychological state in which an individual's personal identity becomes deeply intertwined with a target---be it a group, leader, value, or cause---resulting in porous boundaries between the self and that target. This fusion creates a powerful reciprocal bond where personal agency is channeled into extreme, pro-target behavior, with the individual experiencing a profound “sense of oneness'' that can motivate costly and self-sacrificial actions in defense of the fusion target.}
\\
\\
\texttt{In this task, label the text as:
\begin{itemize}
    \item[-] "low'': Minimal fusion between individual and target identity. Low fusion is marked by a clear separation between the self and the target, so the individual shows little behavioral commitment to the target.
    \item[-] "medium'': Moderate fusion between individual and target identity. Medium fusion reflects a moderate integration where the personal self overlaps with the target enough to inspire occasional support without overwhelming personal autonomy.
    \item[-] "high'': Strong fusion; the individual's identity is almost completely merged with the target's identity. High fusion is characterized by an intense, nearly inseparable merging of identity with the target, driving individuals to engage in extreme, self-sacrificial actions for its sake.
\end{itemize}}

\texttt{Below are three examples:}
\\
\\
\texttt{Example 1 (Lowest Scoring - low):
Text: "\{low\_text\}"
Label: low}
\\
\\
\texttt{Example 2 (Most Middle Scoring - medium):
Text: "\{medium\_text\}"
Label: medium}
\\
\\
\texttt{Example 3 (Highest Scoring - high):
Text: "\{high\_text\}"
Label: high}
\\
\\
\texttt{Now, it's your turn:}
\\
\\
\texttt{Please classify the following text:
Text: "\{sample\_1.iloc[0]['write']\}"
Label: "\{sample\_1.iloc[0]['label']\}"}
\\
\\
\texttt{Please classify the following text:
Text: "\{sample\_2.iloc[0]['write']\}"
Label: "\{sample\_2.iloc[0]['label']\}"}
\\
\\
\texttt{Please classify the following text:
Text: "\{sample\_3.iloc[0]['write']\}"
Label: "\{sample\_3.iloc[0]['label']\}"}
\\
\\
\texttt{Please classify the following text:
Text: "\{text\}"
Label:
}

\subsection{RAG Prompting}

The RAG prompt mostly follows the same pattern as the few-shot approach. The primary difference is that it does not use 3 random samples from the training set. Instead, the text to be classified is converted to a semantic embedding using \texttt{all-mpnet-base-v2}, and then obtains 5 most similar embeddings from the training set as evaluated by FAISS. We use those samples along with their real VIFS scores as examples. 

The RAG prompt is as follows:

\texttt{\textbf{role:} system}
\\
\\
\texttt{\textbf{content:} You are a text classifier that determines the level of identity fusion in a given text. Identity fusion is when an individual's personal identity becomes strongly intertwined with their target's identity.}
\\
\\
\texttt{Based on Swann et al. (2024), identity fusion is a psychological state in which an individual's personal identity becomes deeply intertwined with a target---be it a group, leader, value, or cause---resulting in porous boundaries between the self and that target. This fusion creates a powerful reciprocal bond where personal agency is channeled into extreme, pro-target behavior, with the individual experiencing a profound “sense of oneness'' that can motivate costly and self-sacrificial actions in defense of the fusion target.}
\\
\\
\texttt{In this task, label the text as:
\begin{itemize}
    \item[-] "low'': Minimal fusion between individual and target identity. Low fusion is marked by a clear separation between the self and the target, so the individual shows little behavioral commitment to the target.
    \item[-] "medium'': Moderate fusion between individual and target identity. Medium fusion reflects a moderate
    integration where the personal self overlaps with the target enough to inspire occasional support without overwhelming personal autonomy.
  \item[-] "high'': Strong fusion; the individual's identity is almost completely merged with the target's identity. High fusion is characterized by an intense, nearly inseparable merging of identity with the target, driving individuals to engage in extreme, self-sacrificial actions for its sake.
\end{itemize}}

\texttt{Below are a three examples:}
\\
\\
\texttt{Example 1 (Lowest Scoring - low):
Classify the following text into [low, medium, high]:\\
Text: "\{low\_text\}"\\
Output only the label, nothing else.\\
Label: low}
\\
\\
\texttt{Example 2 (Most Middle Scoring - medium):\\
Classify the following text into [low, medium, high]:\\
Text: "\{medium\_text\}"\\
Output only the label, nothing else.\\
Label: medium}
\\
\\
\texttt{Example 3 (Highest Scoring - high):\\
Classify the following text into [low, medium, high]:\\
Text: "\{medium\_text\}"\\
Output only the label, nothing else.\\
Label: high}
\\
\\
The next part of the prompt is repeated 5 times for the top 5 most similar entries in the training set as returned from FAISS.
\\
\\
\texttt{\textbf{role:} user}
\\
\\
\texttt{\textbf{content:} Classify the following text into [low, medium, high]:\\
Text: "\{RETRIEVED SAMPLE\}"\\
Output only the label, nothing else.\\
Label: }
\\
\\
\texttt{\textbf{role:} assistant\\
\textbf{content:} \{RETRIEVED LABEL\}}
\\
\\
Finally, the model is allowed classify the current text after seeing all retrieved examples.
\\
\\
\texttt{\textbf{role:} user}
\\
\\
\texttt{\textbf{content:} Classify the following text into [low, medium, high]:\\
Text: "\{text\}"\\
Output only the label, nothing else.\\
Label: }

\begin{table*}[!htbp]
\centering
\begin{tabular}{llll}
\hline
\multicolumn{3}{c}{\textbf{Violence Risk Prediction Data}} \\
\hline
\textbf{Author} & \textbf{Description} & \textbf{Label} \\
\hline
Anders Behring Breivik & Manifesto of the Norway attacks, 2011 & VSS \\
Elliot Rodger & Manifesto of the Isla Vista killings, 2014 & VSS \\
Dylann Roof & Manifesto of the Charleston shooting, 2015 & VSS \\
Brenton Tarrant & Manifesto of the Christchurch mosque attacks, 2019 & VSS \\
Stephan Baillet & Manifesto of the Halle synagogue shooting, 2019 & VSS \\
John Earnest & Manifesto of the Poway synagogue attack, 2019 & VSS \\
Patrik Crusius & Manifesto of the El Paso attack, 2019 & VSS \\
Adolf Hitler & Mein Kampf, 1925 & VSS \\
Sayyid Qutb & Milestones, 1964 & VSS \\
Karl Marx \& Friedrich Engels & Manifesto of the Communist Party, 1848 & IE \\
Yusuf al-Qaradawi & The Lawful and Prohibited in Islam, 1960 & IE \\
Fjordman & Defeating Eurabia, 2008 & IE \\
Simone de Beauvoir & The Second Sex, 1949 & M \\
Martin Luther King Jr. & I Have a Dream, 1963 &  M \\
Greta Thunberg & Our House Is on Fire, 2019 & M \\
\hline
\end{tabular}
\caption{Violence Risk Prediction Data \citep{EbnerKavanaghWhitehouse2022}. VSS: Violent Self-Sacrificial; IE: Ideologically Extreme; M: Moderate.}
\label{tab:manifestos}
\end{table*}

\subsection{Use of Scientific Artifacts:}
\label{sec:artifacts}

The identity fusion dataset was introduced in a PNAS Nexus paper published under a CC BY 4.0 license~\citep{AshokkumarPennebaker2022}, with the data provided via the article's supplementary materials. Although the dataset itself does not explicitly include a license statement, PNAS Nexus's policy requires that all supplementary data be publicly available for reproducibility, and the authors indicated in their supplementary material that it was public data~\citep{AshokkumarPennebaker2022}; we therefore understand it falls under the same CC BY 4.0 terms. We note they anonymized the dataset before publication. Consistent with the intended use and terms, we are free to share and adapt this data. We only adapt and reorganize the data during augmentation and train, test, and validation splits. We include the augmented training set as part of our public CLIFS ensemble model (for RAG). In addition, we re-implement the UAI as detailed within the paper, and modify it for our purposes, which is also within its intended use.

The VRI manifesto paper is a CC BY 4.0 licensed paper~\citep{EbnerKavanaghWhitehouse2022}, and the data was noted by its authors as available upon request due to sensitive content. Since the corpus is comprised of only 15 publicly accessible manifestos from prominent individuals, we reconstructed it for our analysis. However, we also do not share this reconstructed dataset publicly, as it contains highly sensitive and harmful material (e.g., from mass killers, terrorists, and extremists). The documents present are specified in Table \ref{tab:manifestos}. As this data accompanies a CC BY 4.0 paper, especially as it is indicated to be made available per request, we are consistent with the intended use as we only reorganize the data (chunking) for analysis. We also partially re-implement their VRI as detailed in their paper \citep{EbnerKavanaghWhitehouse2024b}, which also has a CC BY 4.0 license, and therefore our adaptation falls within its intended use. Our publicly available adaptations remove all categories that are not utilized in our method; this includes removing all categories with harmful language (e.g., with racial slurs).

\onecolumn
\clearpage
\section{Additional Tables \& Figures}
\label{sec:appendixd}

\begin{table*}[!h]
  \centering
  \begin{tabular}{l cccccc}
    \hline
    \multicolumn{7}{c}{\textbf{Overall Performance | Bootstrapped}} \\
    \hline
    \textbf{Model} & \multicolumn{3}{c}{\textbf{Original Data}} & \multicolumn{3}{c}{\textbf{Augmented Data}} \\
                   & $F_1$ & 95\% CI  & CI Width      & $F_1$ & 95\% CI  & 95\% CI Width       \\
    \hline
    \textbf{Majority Vote}    & 0.26 & [0.24, 0.28] & 0.04 & 0.26 & [0.24, 0.28] & 0.04 \\
    \textbf{MB Zero-Shot}     & 0.32 & [0.24, 0.40] & 0.16 & 0.32 & [0.24, 0.40] & 0.16 \\
    \textbf{4o Few-Shot}      & 0.58 & [0.48, 0.66] & 0.18 & 0.43 & [0.33, 0.53] & 0.20 \\
    \textbf{4o RAG}           & 0.57 & [0.48, 0.66] & 0.18 & 0.59 & [0.51, 0.68] & 0.17 \\
    \textbf{r1 RAG}           & 0.61 & [0.52, 0.71] & 0.19 & 0.56 & [0.45, 0.65] & 0.20 \\
    \textbf{SBERT RF}         & 0.59 & [0.48, 0.69] & 0.21 & 0.49 & [0.39, 0.59] & 0.20 \\
    \textbf{ModernBERT}       & 0.49 & [0.38, 0.60] & 0.22 & 0.62 & [0.52, 0.71] & 0.19 \\
    \textbf{CLIFS Ensemble}   & 0.62 & [0.53, 0.71] & 0.18 & \textbf{0.65} & [0.55, 0.74] & 0.19 \\
    \textbf{CLIFS RF}         & 0.55 & [0.46, 0.64] & 0.18 & \textbf{0.65} & [0.56, 0.75] & 0.19 \\
    \textbf{CLIFS XGB}       & - & - & - & - & - & - \\
    \textbf{CLIFS SVM}         & - & - & - & - & - & - \\
    \hline
  \end{tabular}
  \caption{Bootstrapped $F_1$ scores and 95\% confidence intervals for models on Original and Augmented datasets.}
  \label{tab:overall_performance_boot}
\end{table*}

\vspace{10em}

\begin{table*}[!h]
  \centering
  \begin{tabular}{l cccccc}
    \hline
    \multicolumn{7}{c}{\textbf{Human Comparison | Bootstrapped}} \\
    \hline
    \textbf{Model} 
      & \multicolumn{3}{c}{\textbf{Original Data}} 
      & \multicolumn{3}{c}{\textbf{Augmented Data}} \\
             & $F_1$ & 95\% CI & 95\% CI Width
      & $F_1$ & 95\% CI & CI Width \\
    \hline
    \textbf{Human}             & 0.46 & [0.34, 0.57]  & 0.23 & 0.46 & [0.34, 0.57] & 0.23 \\
    \textbf{Majority Vote}     & 0.25 & [0.22, 0.27]  & 0.05 & 0.25 & [0.22, 0.27] & 0.05 \\
    \textbf{MB Zero-Shot}      & 0.39 & [0.30, 0.50]  & 0.20 & 0.39 & [0.30, 0.50] & 0.20 \\
    \textbf{4o Few-Shot}       & 0.37 & [0.30, 0.43]  & 0.13 & 0.54 & [0.43, 0.66] & 0.23 \\
    \textbf{4o RAG}            & 0.54 & [0.43, 0.64]  & 0.21 & \textbf{0.59} & [0.49, 0.69] & 0.20 \\
    \textbf{r1 RAG}            & \textbf{0.59} & [0.48, 0.70] & 0.22 & 0.58 & [0.47, 0.69] & 0.22 \\
    \textbf{SBERT RF}          & 0.43 & [0.36, 0.50]  & 0.14 & 0.43 & [0.35, 0.50] & 0.15 \\
    \textbf{ModernBERT}        & 0.40 & [0.33, 0.48]  & 0.15 & 0.52 & [0.42, 0.63] & 0.21 \\
    \textbf{CLIFS Ensemble}    & 0.52 & [0.40, 0.63]  & 0.23 & 0.56 & [0.45, 0.67] & 0.22 \\
    \textbf{CLIFS RF}          & 0.55 & [0.44, 0.67]  & 0.23 & 0.51 & [0.42, 0.62] & 0.20  \\
    \textbf{CLIFS XGB}         & - & - & - & - & - & - \\
    \textbf{CLIFS SVM}         & - & - & - & - & - & - \\
    \hline
  \end{tabular}
  \caption{Bootstrapped $F_1$ scores and 95\% confidence intervals for the human-comparison benchmark on Original and Augmented datasets.}
  \label{tab:human_comparison_boot}
\end{table*}

\begin{table*}
  \centering
  \begin{tabular}[!h]{lcccccccc}
  \hline
    \multicolumn{9}{c}{\textbf{Overall Per Class Performance}} \\
    \hline
    \textbf{Model}
      & \multicolumn{4}{c}{\textbf{Original}}
      & \multicolumn{4}{c}{\textbf{Augmented}} \\
    \cline{2-5}\cline{6-9}
      & \textbf{$F_1$}
      & \textbf{Low}
      & \textbf{Medium}
      & \textbf{High}
      & \textbf{$F_1$}
      & \textbf{Low}
      & \textbf{Medium}
      & \textbf{High} \\
    \hline
    \textbf{Majority Vote}
      & 0.26 & 0.00 & 0.78 & 0.00
      & 0.26 & 0.00 & 0.78 & 0.00 \\

    \textbf{MB Zero-Shot}
      & 0.32 & 0.40 & 0.32 & 0.24
      & 0.32 & 0.40 & 0.32 & 0.24 \\

    \textbf{4o Few-Shot}
      & 0.58 & 0.53 & 0.62 & 0.59
      & 0.43 & 0.43 & 0.48 & 0.38 \\

    \textbf{4o RAG}
      & 0.57 & 0.50 & 0.60 & 0.62
      & 0.60 & 0.53 & 0.65 & 0.60 \\

    \textbf{r1 RAG}
      & 0.62 & 0.57 & 0.72 & 0.57
      & 0.56 & 0.56 & 0.70 & 0.42 \\

    \textbf{SBERT RF}
      & 0.59 & 0.54 & 0.78 & 0.46
      & 0.50 & 0.41 & \textbf{0.79} & 0.29 \\

    \textbf{ModernBERT}
      & 0.49 & 0.25 & 0.78 & 0.44
      & 0.62 & 0.56 & 0.75  & 0.56 \\

    \textbf{CLIFS Ensemble}
      & 0.63 & 0.58 & 0.69 & 0.61
      & \textbf{0.66} & 0.59 & 0.73  & \textbf{0.65} \\

    \textbf{CLIFS RF}
      & 0.55 & 0.51 & 0.67 & 0.49
      & \textbf{0.66} & \textbf{0.62} & 0.78 & 0.58 \\

    \textbf{CLIFS XGB}
      & 0.54 & 0.43 & 0.77 & 0.43
      & 0.58 & 0.52 & 0.76 & 0.45 \\

    \textbf{CLIFS SVM}
      & 0.58 & 0.53 & 0.65 & 0.58
      & \textbf{0.66} & 0.59 & 0.78 & 0.63 \\

    \hline
  \end{tabular}
  \caption{Overall and per‐class $F_1$ scores for each model trained on the original or augmented data.}
  \label{tab:performance_per_class}
\end{table*}

\begin{table*}
  \centering
  \begin{tabular}[!h]{lcccccccc}
    \hline
    \multicolumn{9}{c}{\textbf{Human Comparison Per Class Performance}} \\
    \hline
    \textbf{Model}
      & \multicolumn{4}{c}{\textbf{Original}}
      & \multicolumn{4}{c}{\textbf{Augmented}} \\
    \cline{2-5}\cline{6-9}
      & \textbf{$F_1$}
      & \textbf{Low}
      & \textbf{Medium}
      & \textbf{High}
      & \textbf{$F_1$}
      & \textbf{Low}
      & \textbf{Medium}
      & \textbf{High} \\
    \hline
    \textbf{Human}
      & 0.46 & 0.32 & 0.70 & 0.36
      & 0.46 & 0.32 & 0.70 & 0.36 \\

    \textbf{Majority Vote}
      & 0.25 & 0.00 & 0.74 & 0.00
      & 0.25 & 0.00 & 0.74 & 0.00 \\

    \textbf{MB Zero-Shot}
      & 0.39 & 0.54 & 0.48 & 0.17
      & 0.39 & 0.54 & 0.48 & 0.17 \\

    \textbf{4o Few-Shot}
      & 0.37 & 0.62 & 0.49 & 0.00
      & 0.54 & 0.68 & 0.71 & 0.24 \\

    \textbf{4o RAG}
      & 0.54 & 0.70 & 0.59 & 0.33
      & \textbf{0.59} & 0.69 & 0.67 & 0.41 \\

    \textbf{r1 RAG}
      & \textbf{0.59} & \textbf{0.71} & 0.71 & 0.35
      & \textbf{0.59} & 0.63 & 0.67 & \textbf{0.46} \\

    \textbf{SBERT RF}
      & 0.43 & 0.57 & 0.71 & 0.00
      & 0.43 & 0.55 & 0.73 & 0.00 \\

    \textbf{ModernBERT}
      & 0.40 & 0.45 & 0.76 & 0.00
      & 0.52 & 0.64 & \textbf{0.78} & 0.14 \\

    \textbf{CLIFS Ensemble}
      & 0.52 & 0.67 & 0.62 & 0.27
      & 0.56 & 0.63 & 0.64 & 0.40 \\

    \textbf{CLIFS RF}
      & 0.56 & 0.69 & 0.72 & 0.27
      & 0.51 & 0.67 & 0.72 & 0.14 \\

    \textbf{CLIFS XGB}
      & 0.43 & 0.37 & 0.76 & 0.15
      & 0.55 & 0.61 & 0.77 & 0.27 \\

    \textbf{CLIFS SVM}
      & 0.52 & 0.63 & \textbf{0.78} & 0.14
      & 0.53 & 0.47 & 0.76 & 0.35 \\

    \hline
  \end{tabular}
  \caption{Overall and per‐class $F_1$ scores for each model on the human‐comparison benchmark, trained on either the original or augmented data.}
  \label{tab:human_comparison_per_class}
\end{table*}

\begin{table*}[ht]
\centering
\renewcommand{\arraystretch}{1.2} %

\begin{tabular}{p{0.12\linewidth} @{\hskip 6pt} p{0.33\linewidth} @{\hskip 12pt} p{0.12\linewidth} @{\hskip 6pt} p{0.33\linewidth}}
\hline
\textbf{Acronym /} & \textbf{Definition} & \textbf{Acronym /} & \textbf{Definition} \\
\textbf{Symbol} & & \textbf{Symbol} & \\
\hline
\textsc{cift} & Comprehensive Identity Fusion Theory & CLIFS & Cognitive Linguistic Identity Fusion Score \\
\hline
DIFI & Dynamic Identity Fusion Index & VIFS & Verbal Identity Fusion Scale \\
\hline
UAI & Unquestioning Affiliation Index & VRI & Violence Risk Index \\
\hline
nUAI & naïve Unquestioning Affiliation Index & RTT & Round-Trip Translation \\
\hline
GenAI & Generative AI & SVM & Support Vector Machine \\
\hline
XGBoost and XGB & Extreme Gradient Boosting & RF & Random Forest \\
\hline
MAE & Mean Absolute Error & GI & Gini Importance \\
\hline
CI & Confidence Interval & RAG & Retrieval-Augmented Generation \\
\hline
Masked-LM & Masked Language Model & $r_s$ & Spearman correlation \\
\hline
$f_{(I,T)}$ & Fusion Proximity & $K_f$ & Fictive Kinship \\
\hline
$S_{I \to T}$ & Directional Proximity (Identity $\to$ Target) & $S_{T \to I}$ & Directional Proximity (Target $\to$ Identity) \\
\hline
$T$ & Fusion Target vocabulary & $I$ & Identity vocabulary; First-person singular pronouns \\
\hline
$K$ & Fictive Kinship vocabulary & $C_m$ & Surrounding context for masked word $m$ \\
\hline
$M_y$ & Total number of masked positions when masking vocabulary $y$ within a given text & $\mathcal{V}_x$ & Vocabulary for category $x$ \\
\hline
$w_v$ & Current word from vocabulary $\mathcal{V}_x$ ($x \in \{I, T, K\}$) replacing word $m$ & $m$ & Current word from vocabulary $y$ being replaced by each word in vocabulary $x$ \\
\hline
\end{tabular}
\caption{Summary of acronyms and symbols used in this paper.}
\label{tab:acronyms_symbols}
\end{table*}

\twocolumn

\begin{figure*}[!h]
  \includegraphics[width=1\textwidth]{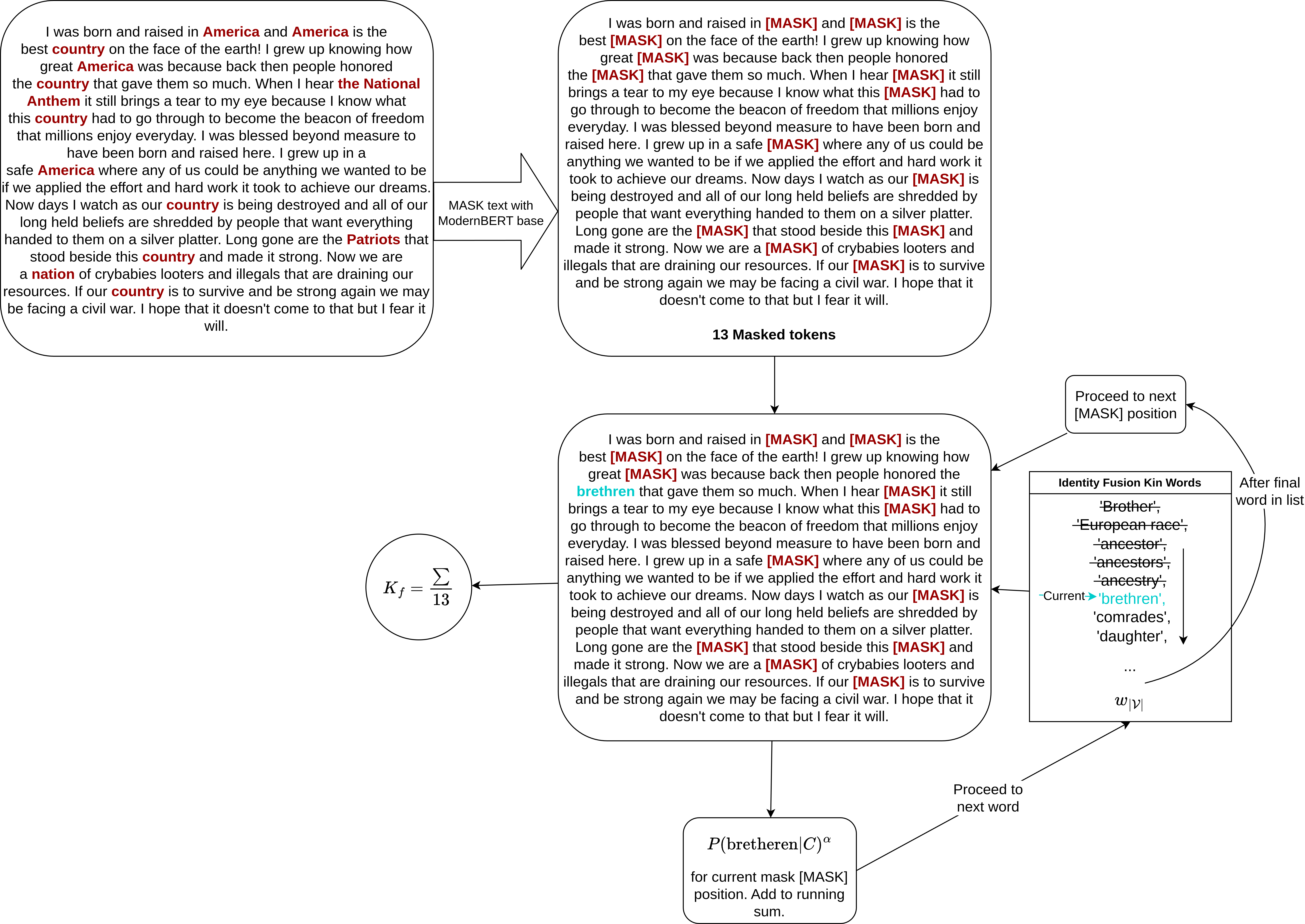}
  \caption{Example calculation of ModernBERT identity fusion scores, specifically $K_f$.}
  \label{fig:mbif_example}
\end{figure*}

\begin{figure*}[!h]
  \centering
  \includegraphics[width=.8\textwidth]{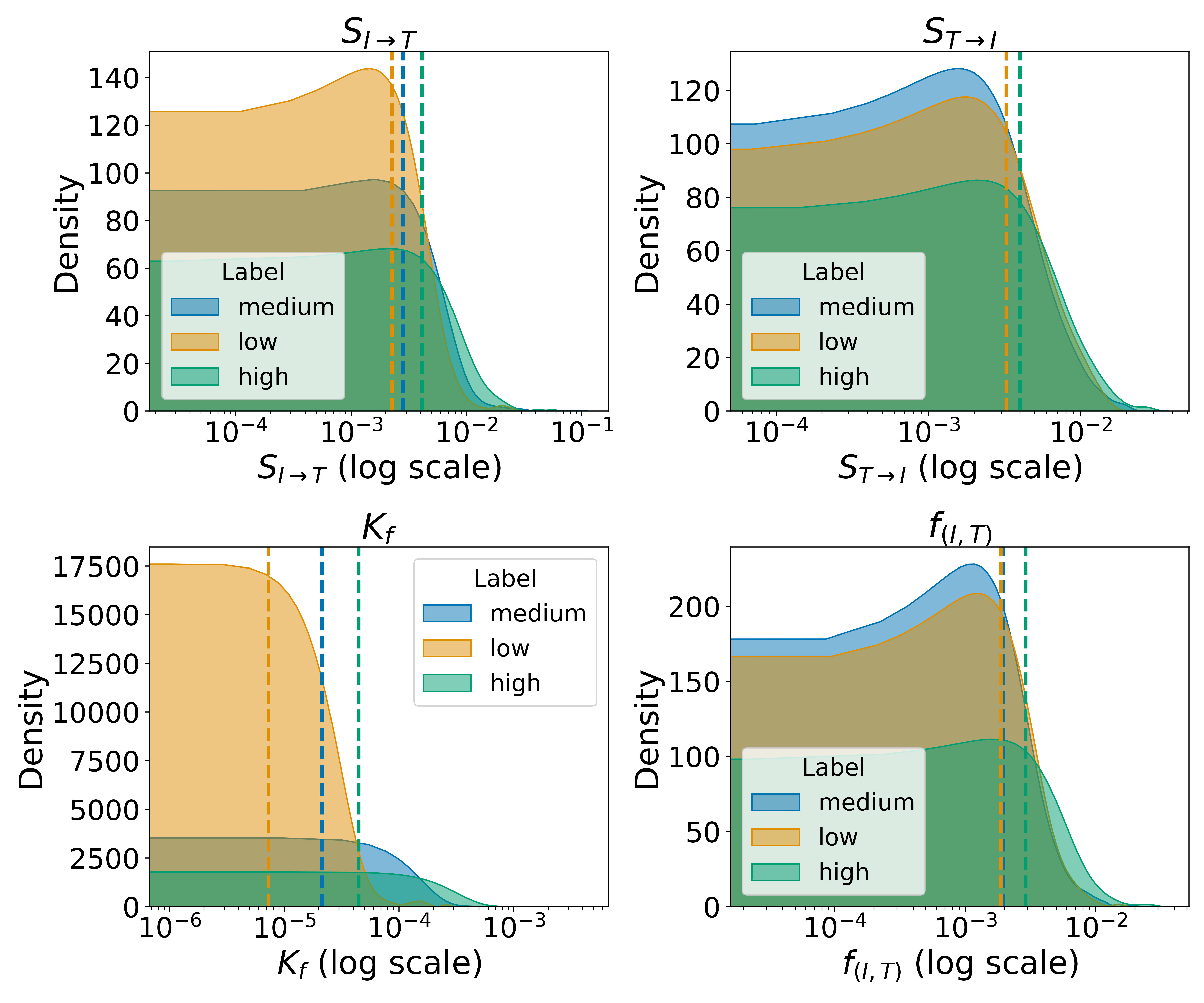}
  \caption{%
    Kernel Density Estimation (KDE) plots of the distributions for the same metrics, separated by true label (means shown as dashed lines; x‐axis log‐scaled).%
  }
  \label{fig:fe_two_panels}
\end{figure*}

\begin{figure}[!h]
  \centering
  \includegraphics[width=.47\textwidth]{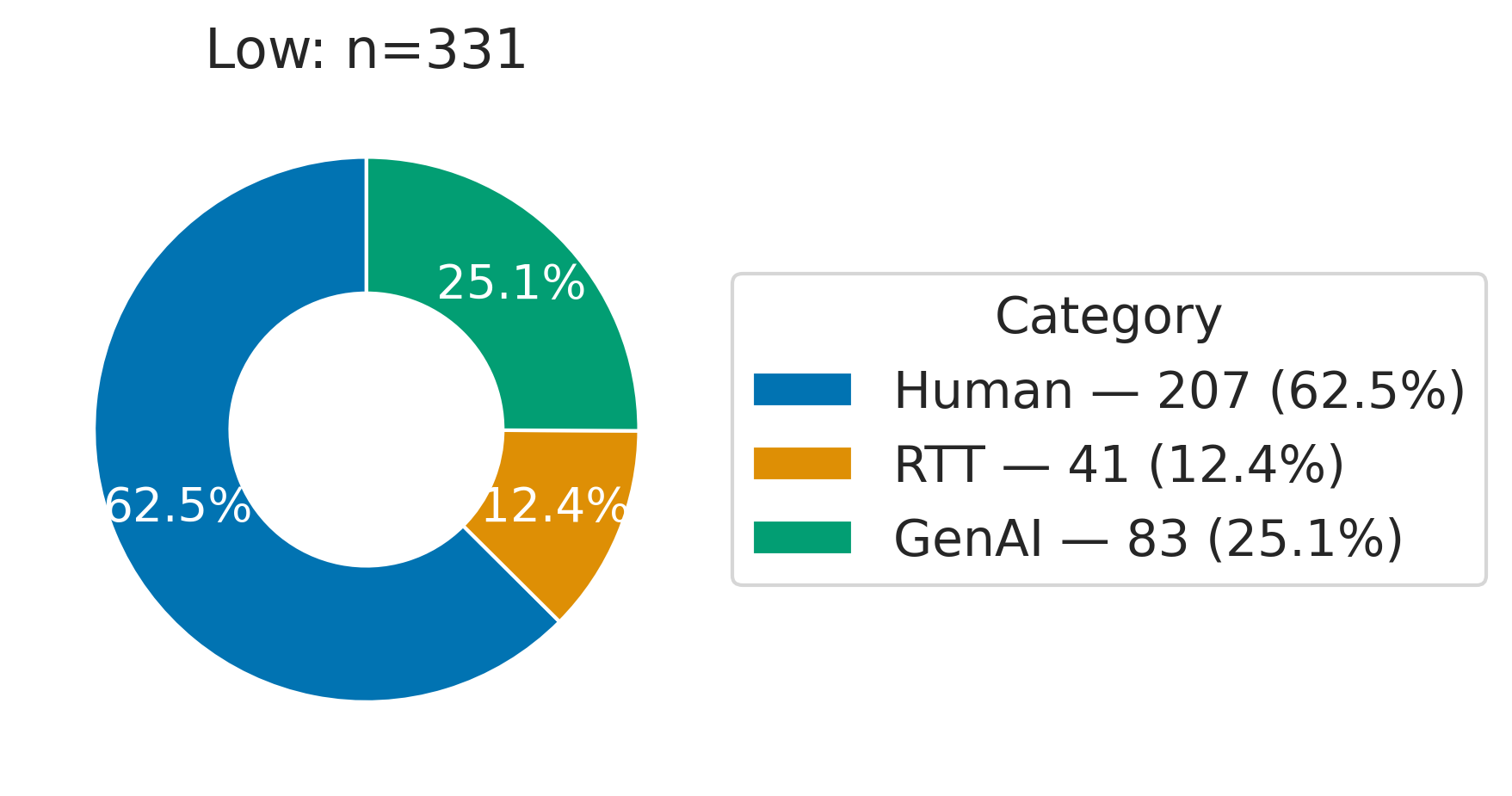}
  \includegraphics[width=.47\textwidth]{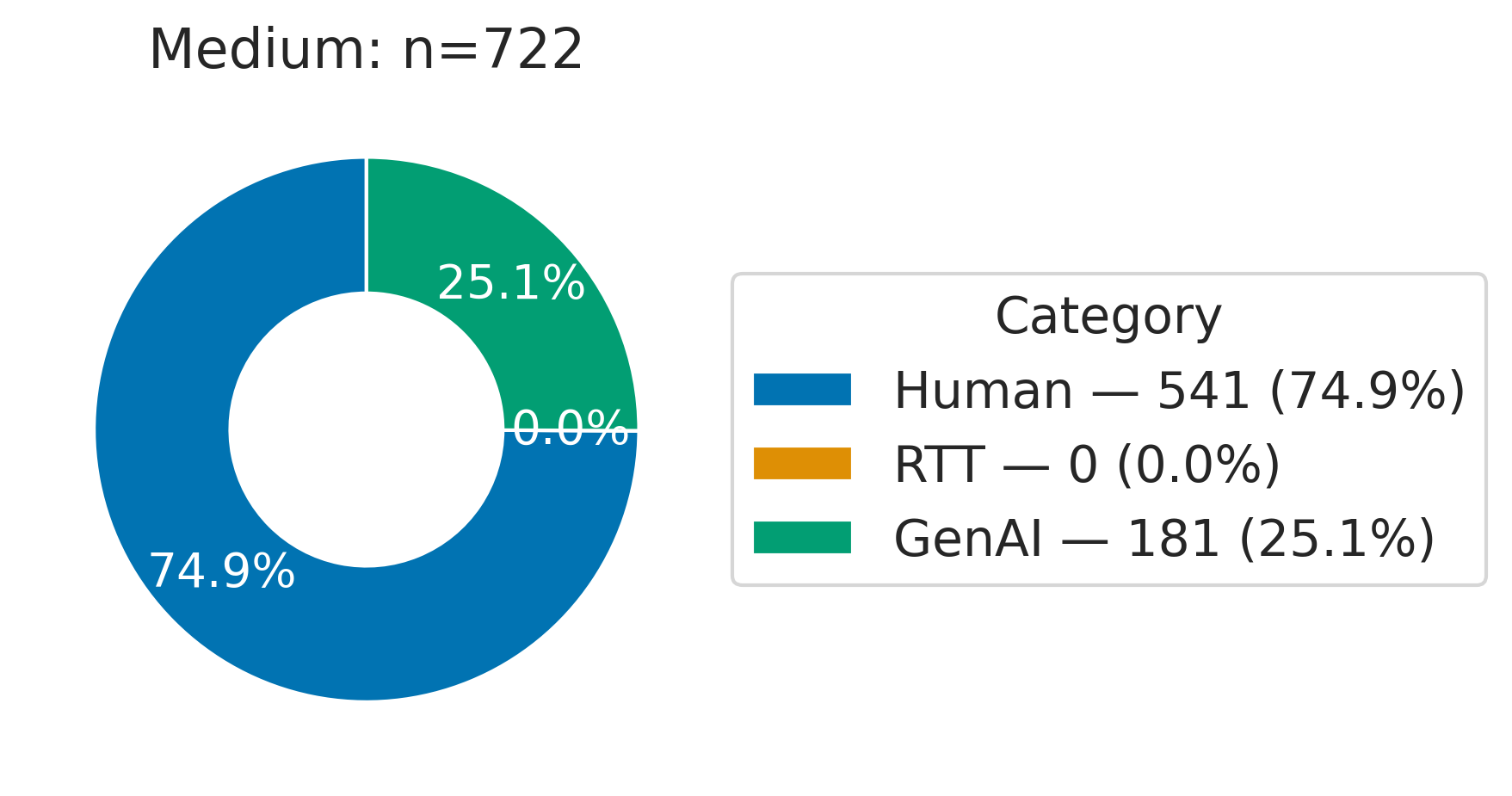}
  \includegraphics[width=.47\textwidth]{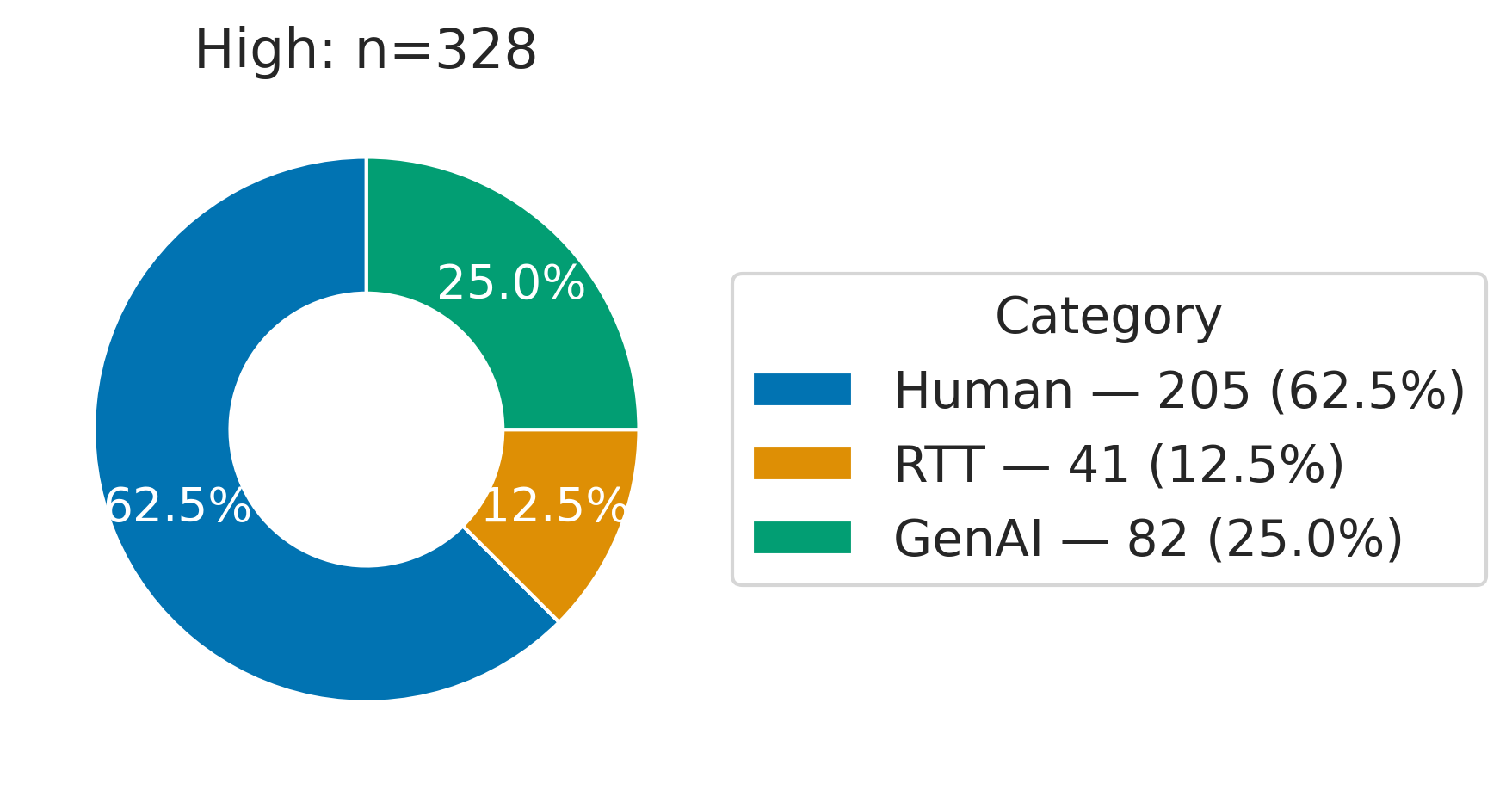}
  \caption{%
    Distribution of human, round-trip translation, and generative AI data after data augmentation.%
  }
  \label{fig:humaidist}
\end{figure}

\begin{figure}[!h]
  \centering
  \includegraphics[width=.482\textwidth]{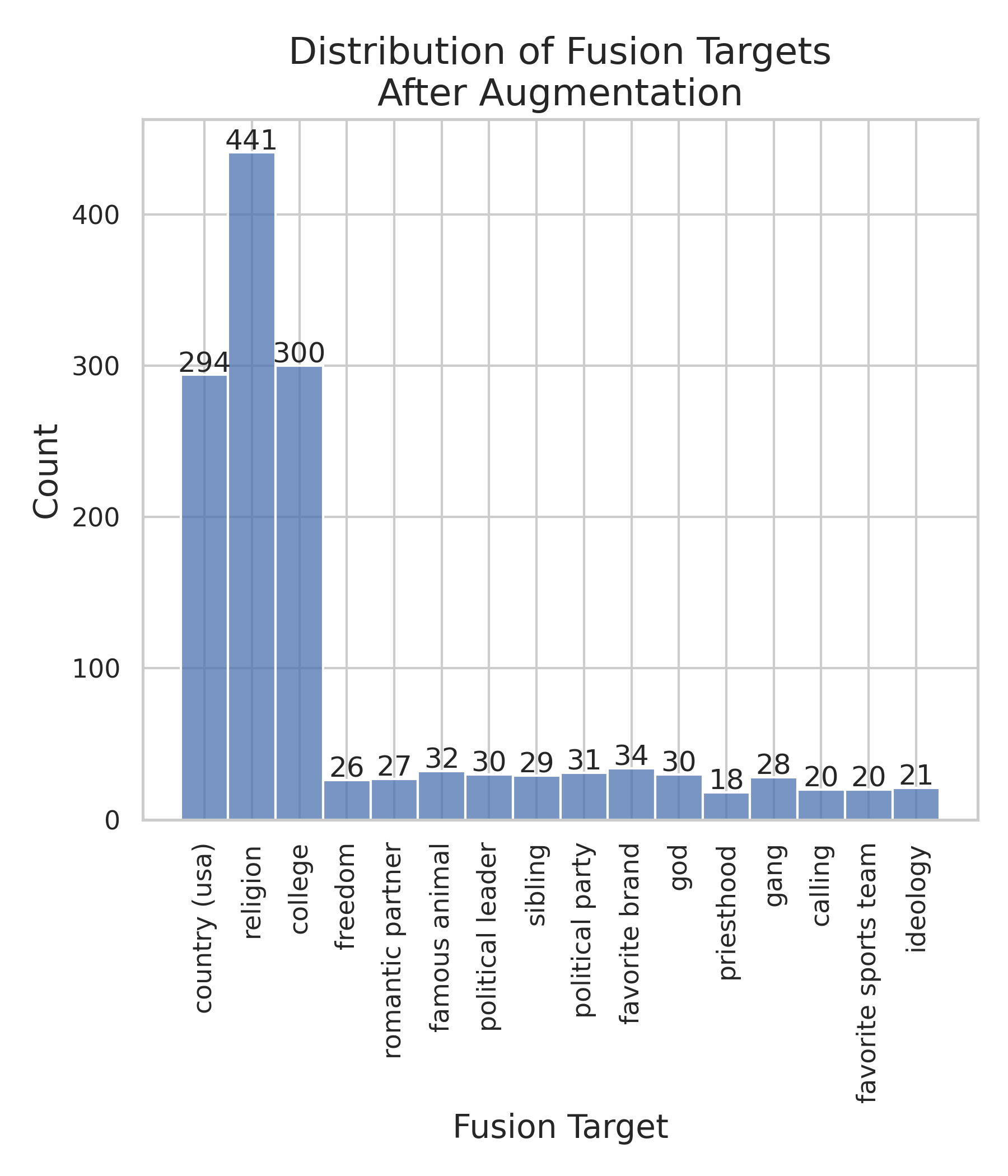}
  \caption{%
    Distribution of fusion-targets after data augmentation.%
  }
  \label{fig:group_dist}
\end{figure}

\begin{figure}
    \includegraphics[width=.482\textwidth]{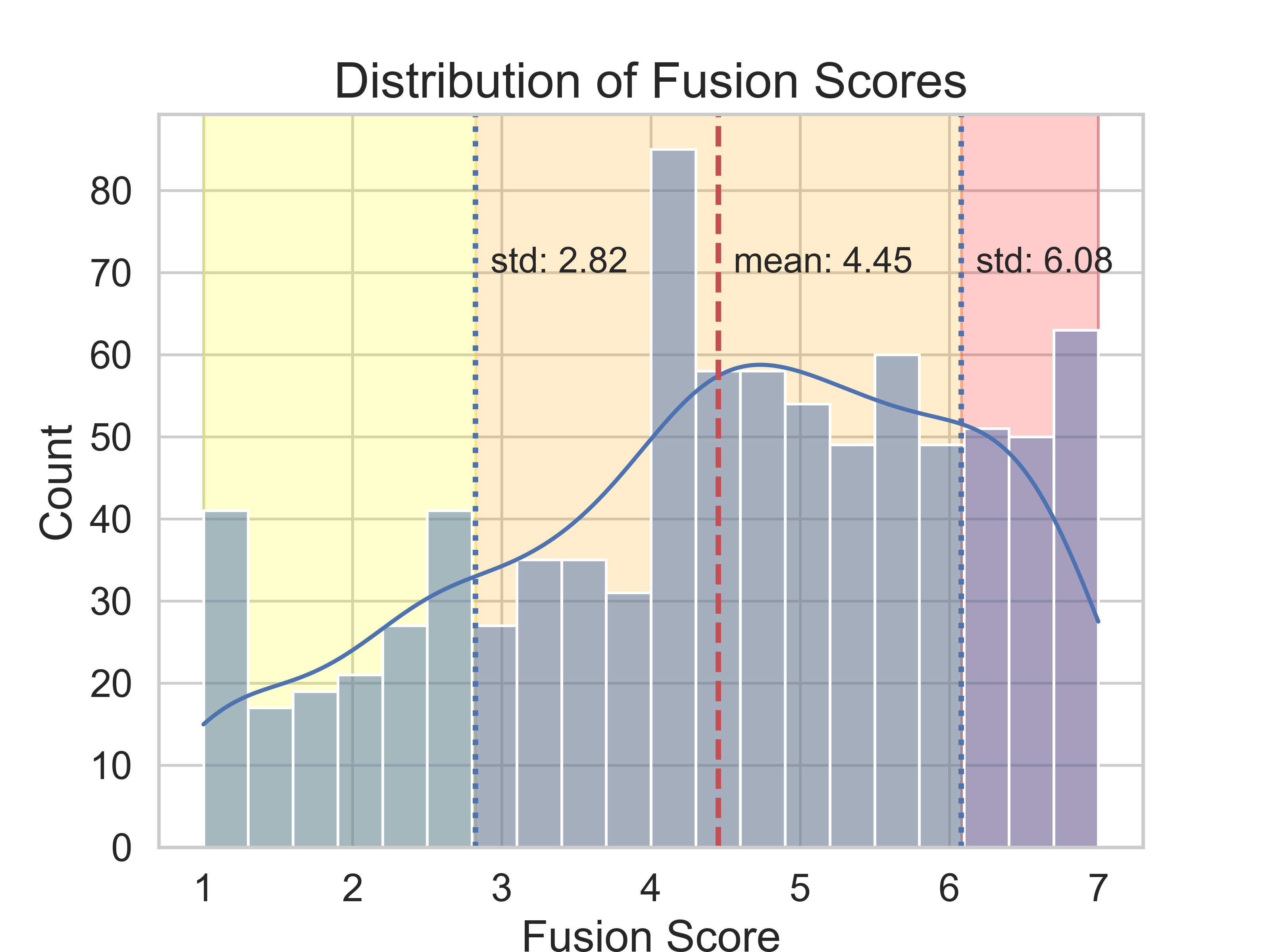}
    \caption{Fusion score distribution of raw data with class discretization. All scores beyond one standard deviation away from the mean Identity Fusion score are classified as ``low'' or ``high;'' reflecting whether they fall below or above the mean.}
    \label{fig:classdiscrete}
\end{figure}

\begin{figure}[t]
  \centering
  \includegraphics[width=0.482\textwidth]{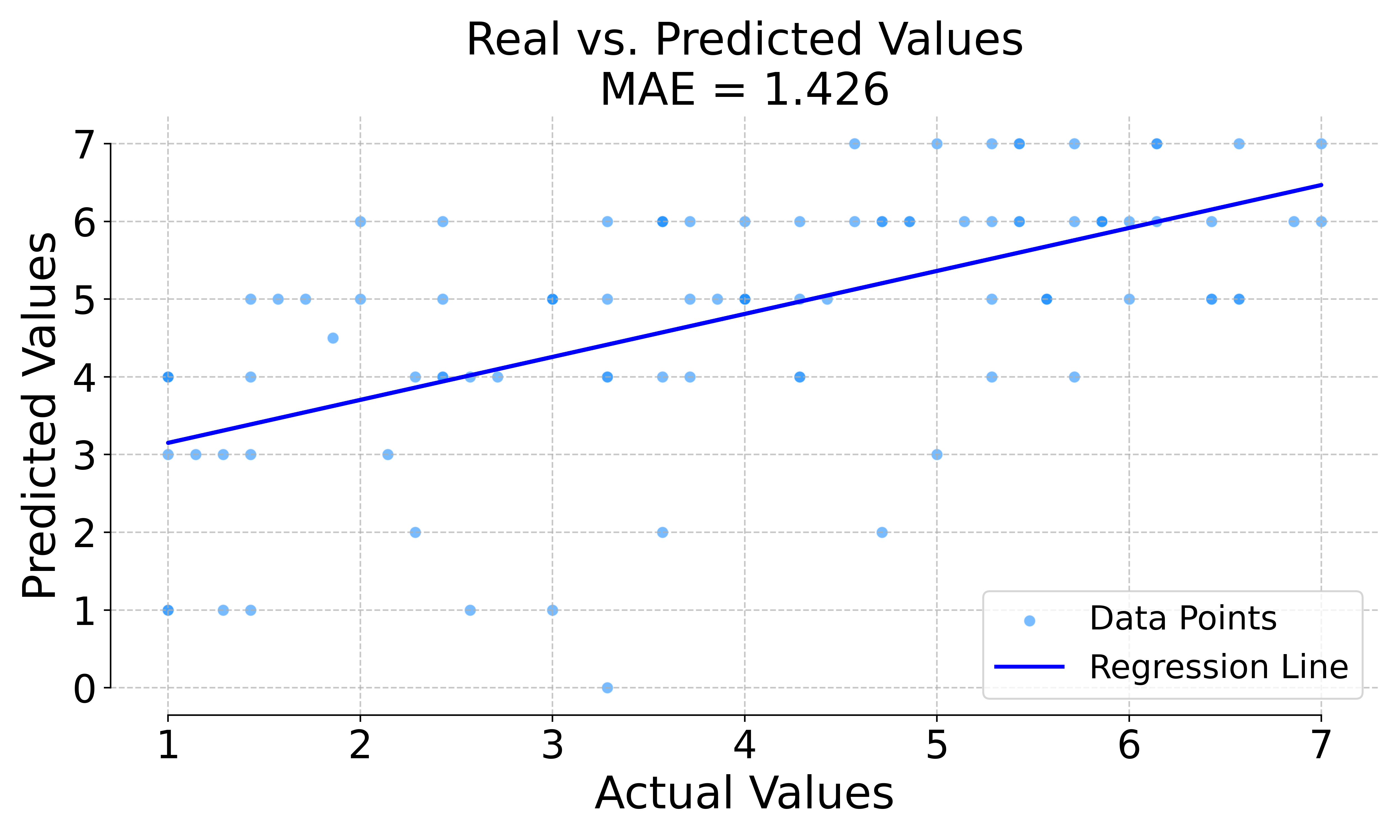}
  \includegraphics[width=0.482\textwidth]{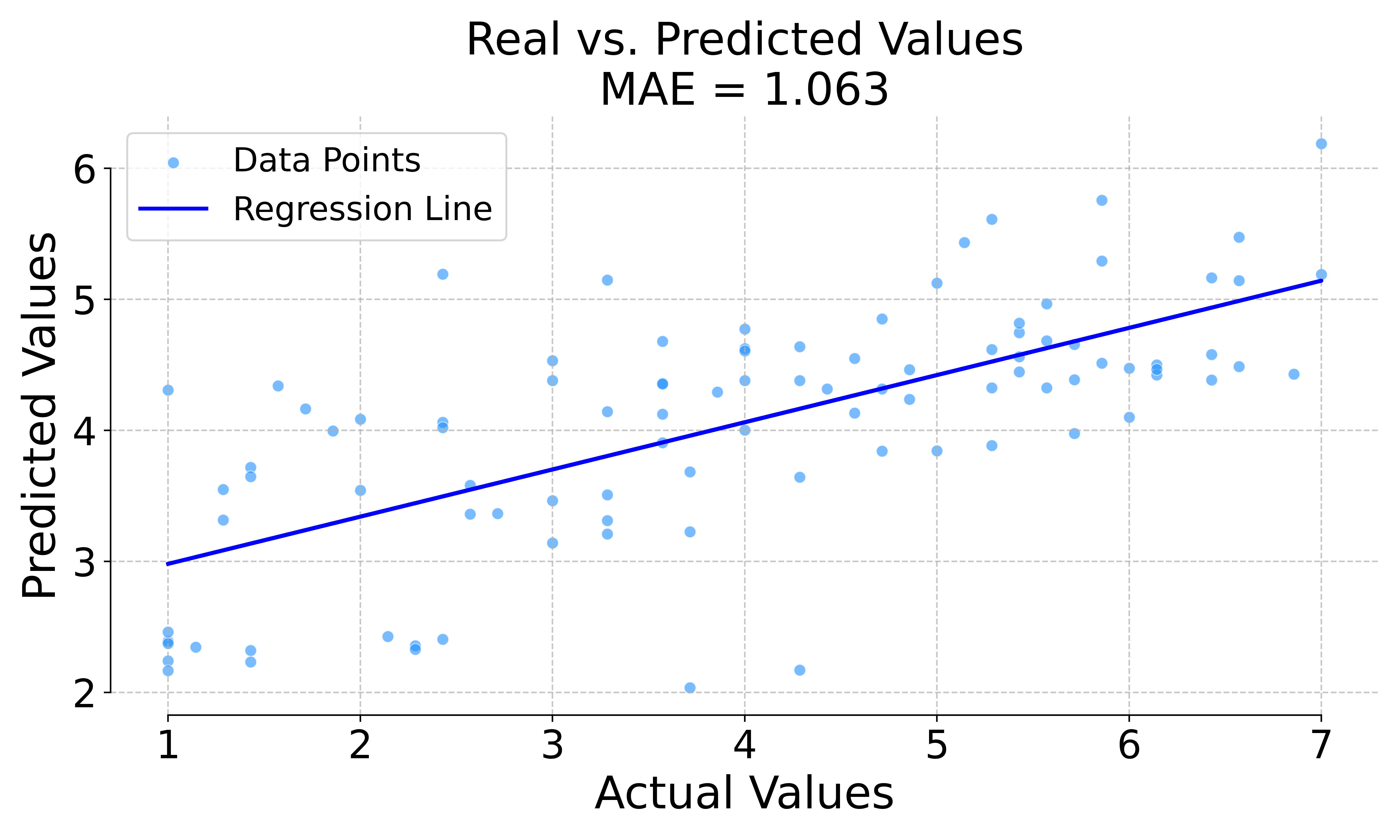}
  \caption {\textbf{Top:} Human identity fusion ratings plotted against the actual identity fusion values as measured from VIFS. $\textnormal{MAE} = 1.426$, $r_s = 0.628$, $p \ll 0.001$. \textbf{Bottom:} The Random Forest regression model trained on augmented data. Tested on human comparison test set. Also plotted against true VIFS values. $\textnormal{MAE} = 1.063$, $r_s = 0.69$, $p \ll 0.001$.}
    \label{fig:regression_human_comparison}
\end{figure}

\begin{figure*}[!htbp]
  \includegraphics[width=1\textwidth]{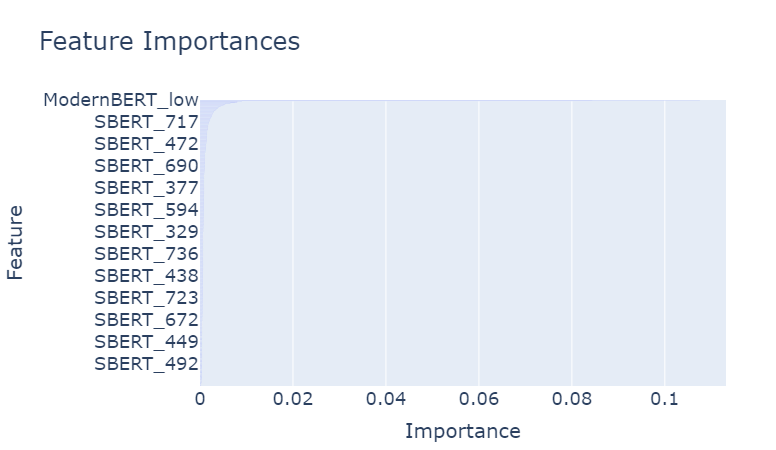}
  \caption{Feature importances (Gini Importance) from the Random Forest model. Each bar shows the relative importance of a feature, as returned by scikit‑learn's feature\_importances\_ attribute. This reflects the mean normalized sum of Gini impurity reductions for that feature across all trees. Higher values indicate greater contributions to reducing impurity, and thus greater influence on the model's performance.}
  \label{fig:all_features_importance}
\end{figure*}

\begin{figure*}[t]
  \includegraphics[width=0.498\linewidth]{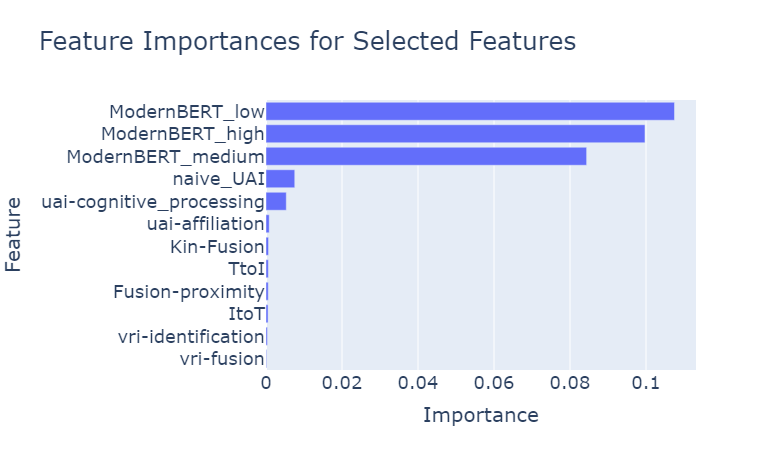} \hfill
  \includegraphics[width=0.498\linewidth]{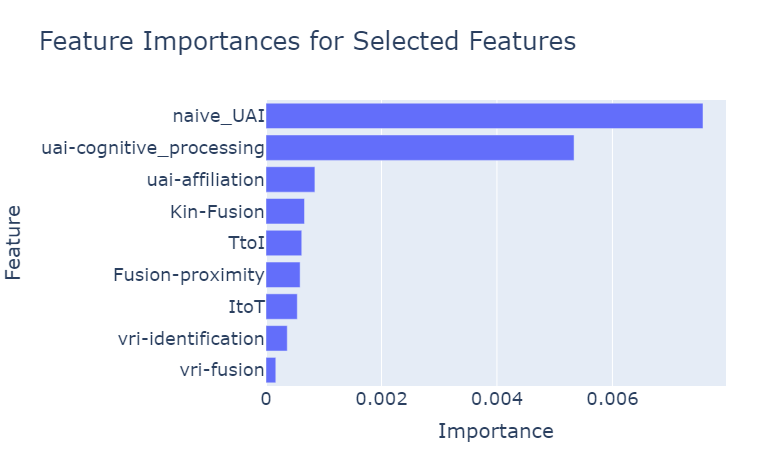}
  \caption {\textbf{Left:} Feature importances for all features used in CLIFS except for SBERT embedding features. \textbf{Right:} The feature importances for all interpretable features from CLIFS.}
    \label{fig:filtered_features_importance}
\end{figure*}

\begin{figure*}[!htbp]
  \includegraphics[width=1\textwidth]{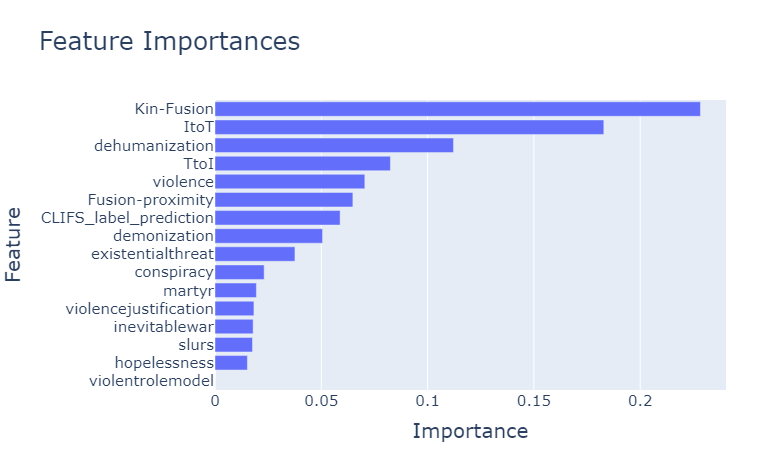}
  \caption{Feature importances for all features in the CLIFS-VRI random forest.}
  \label{fig:vri_features_importance}
\end{figure*}

\clearpage
\section{VIFS \& Sets}

\subsection{The 7-item Verbal Identity Fusion Scale Questions}
\label{sec:appendixa7}

\begin{enumerate}
    \item My [target] is me.
    \item I am one with my [target].
    \item I feel immersed in my [target].
    \item I have a deep emotional bond with my [target].
    \item I am strong because of my [target].
    \item I'll do for my [target] more than any of the other [group members/etc.] would do.
    \item I make my [target] strong.
\end{enumerate}

As mentioned above, the country-target participants only answered the following subset.

\begin{enumerate}
    \item I am one with my [target].
    \item I have a deep emotional bond with my [target].
    \item I am strong because of my [target].
    \item I make my [target] strong.
\end{enumerate}

\subsection{Sets: I, T, K}
\label{sec:seeds}

\begin{enumerate}
    \item \textbf{I:} i, me, my, mine, myself.
    \item \textbf{T:}
        \begin{enumerate}
            \item \textbf{First Person Plural Pronouns:} we, us, our, ours, ourselves
            \item \textbf{Specific | Parameter:} religion, religious, church, god, college, university, school, usa, country, America, (seed set)
            \item \textbf{Generic | Not a Parameter:} team, class, club, society, squad, gang, band, crew (generic collective set)
        \end{enumerate}
    \item \textbf{K:} Brother, sister, family, motherland, our blood, fatherland, sons, daughters, kin, my people, my race, our people, European race, ancestry, ancestor, descendant, fellow, brethren, comrades (seed set)
\end{enumerate}

\clearpage
\section{Model Details \& Resources}
\label{sec:appendixparamrec}

\subsection{LLM Parameter Size}

\begin{enumerate}
    \item SBERT 
        \begin{enumerate}
            \item \texttt{all-mpnet-base-v2}
            \begin{enumerate}
                \item 109M parameters
            \end{enumerate}
        \end{enumerate}
    \item Answer.AI
        \begin{enumerate}
            \item \texttt{ModernBERT-base}
                \begin{enumerate}
                    \item 149M parameters
                \end{enumerate}
        \end{enumerate}
    \item DeepSeek R1
        \begin{enumerate}
            \item \texttt{deepseek-reasoner}
                \begin{enumerate}
                    \item 685B parameters
                \end{enumerate}
        \end{enumerate}
    \item OpenAI GPT-4o
        \begin{enumerate}
            \item \texttt{gpt-4o}
                \begin{enumerate}
                    \item 200B parameters
                \end{enumerate}
        \end{enumerate}
    \item Helsinki-NLP
        \begin{enumerate}
            \item \texttt{opus-mt}
                \begin{enumerate}
                    \item 77.9M parameters
                \end{enumerate}
        \end{enumerate}
    \item Facebook
        \begin{enumerate}
            \item \texttt{wmt19}
                \begin{enumerate}
                    \item 270M parameters
                \end{enumerate}
        \end{enumerate}
\end{enumerate}

\subsection{Hyperparameters}

Final hyperparameters for all \textbf{classifiers} performing \textbf{identity fusion prediction}:

\begin{enumerate}
    \item \textbf{CLIFS Random Forest:}
        \begin{enumerate}
            \item \textbf{Overall:}
                \begin{enumerate}
                    \item \textbf{Raw Data:}
                        \begin{enumerate}
                            \item classifier\_\_max\_depth: None
                            \item classifier\_\_min\_samples\_leaf: 5
                            \item classifier\_\_min\_samples\_split: 20
                            \item classifier\_\_n\_estimators: 300
                            \item scaler: passthrough
                        \end{enumerate}
                    \item \textbf{Augmented Data:}
                        \begin{enumerate}
                            \item classifier\_\_max\_depth: 20
                            \item classifier\_\_min\_samples\_leaf: 2
                            \item classifier\_\_min\_samples\_split: 20
                            \item classifier\_\_n\_estimators: 400
                            \item scaler: RobustScaler()
                        \end{enumerate}
                \end{enumerate}
            \item \textbf{Human Comparison:}
                \begin{enumerate}
                    \item \textbf{Raw Data:}
                        \begin{enumerate}
                            \item classifier\_\_max\_depth: None
                            \item classifier\_\_min\_samples\_leaf: 5
                            \item classifier\_\_min\_samples\_split: 20
                            \item classifier\_\_n\_estimators: 50
                            \item scaler: passthrough
                        \end{enumerate}
                    \item \textbf{Augmented Data:}
                        \begin{enumerate}
                            \item classifier\_\_max\_depth: None
                            \item classifier\_\_min\_samples\_leaf: 5
                            \item classifier\_\_min\_samples\_split: 2
                            \item classifier\_\_n\_estimators: 200
                            \item scaler: passthrough 
                        \end{enumerate}
                \end{enumerate}
        \end{enumerate}
    \item \textbf{SBERT Random Forest:}
        \begin{enumerate}
            \item \textbf{Overall:}
                \begin{enumerate}
                    \item \textbf{Raw Data:}
                        \begin{enumerate}
                            \item classifier\_\_max\_depth: None
                            \item classifier\_\_min\_samples\_leaf: 10
                            \item classifier\_\_min\_samples\_split: 2
                            \item classifier\_\_n\_estimators: 300
                            \item scaler: passthrough
                        \end{enumerate}
                    \item \textbf{Augmented Data:}
                        \begin{enumerate}
                            \item classifier\_\_max\_depth: 20
                            \item classifier\_\_min\_samples\_leaf: 1
                            \item classifier\_\_min\_samples\_split: 20
                            \item classifier\_\_n\_estimators: 200
                            \item scaler: passthrough
                        \end{enumerate}
                \end{enumerate}
            \item \textbf{Human Comparison:}
                \begin{enumerate}
                    \item \textbf{Raw Data:}
                        \begin{enumerate}
                            \item classifier\_\_max\_depth: None
                            \item classifier\_\_min\_samples\_leaf: 10
                            \item classifier\_\_min\_samples\_split: 2
                            \item classifier\_\_n\_estimators: 100
                            \item scaler: passthrough
                        \end{enumerate}
                    \item \textbf{Augmented Data:}
                        \begin{enumerate}
                            \item classifier\_\_max\_depth: None
                            \item classifier\_\_min\_samples\_leaf: 5
                            \item classifier\_\_min\_samples\_split: 2
                            \item classifier\_\_n\_estimators: 400
                            \item scaler: passthrough 
                        \end{enumerate}
                \end{enumerate}
        \end{enumerate}
    \item \textbf{Fine-Tuned ModernBERT:}
        \begin{enumerate}
            \item \textbf{Overall:}
                \begin{enumerate}
                    \item \textbf{Raw Data:}
                        \begin{enumerate}
                            \item learning\_rate: 1.447634258437072e-05
                            \item per\_device\_train\_batch\_size: 32
                            \item per\_device\_eval\_batch\_size: 32
                            \item weight\_decay: 0.002741795210253083
                            \item num\_train\_epochs: 4
                            \item warmup\_ratio: 0.2984258360785583
                            \item lr\_scheduler\_type: polynomial
                        \end{enumerate}
                    \item \textbf{Augmented Data:}
                        \begin{enumerate}
                            \item learning\_rate: 0.00019174112428857004
                            \item per\_device\_train\_batch\_size: 32
                            \item per\_device\_eval\_batch\_size: 64
                            \item weight\_decay: 0.00595353861040398
                            \item num\_train\_epochs: 3
                            \item warmup\_ratio: 0.07542637670184059
                            \item lr\_scheduler\_type: polynomial
                        \end{enumerate}
                \end{enumerate}
            \item \textbf{Human Comparison:}
                \begin{enumerate}
                    \item \textbf{Raw Data:}
                        \begin{enumerate}
                            \item learning\_rate: 7.459295575723428e-05
                            \item per\_device\_train\_batch\_size: 16
                            \item per\_device\_eval\_batch\_size: 32
                            \item weight\_decay: 0.0010037021913674917
                            \item num\_train\_epochs: 4
                            \item warmup\_ratio: 0.25014457922189737
                            \item lr\_scheduler\_type: cosine
                        \end{enumerate}
                    \item \textbf{Augmented Data:}
                        \begin{enumerate}
                            \item learning\_rate: 0.00011843171658742821
                            \item per\_device\_train\_batch\_size: 16
                            \item per\_device\_eval\_batch\_size: 64
                            \item weight\_decay: 0.001181105691906098
                            \item num\_train\_epochs: 2
                            \item warmup\_ratio: 0.13989389333316193
                            \item lr\_scheduler\_type: polynomial
                        \end{enumerate}
                \end{enumerate}
        \end{enumerate}
    \item \textbf{CLIFS Extreme Gradient Boosting:}
        \begin{enumerate}
            \item \textbf{Overall:}
                \begin{enumerate}
                    \item \textbf{Raw Data:}
                        \begin{enumerate}
                            \item classifier\_\_subsample: 1.0
                            \item classifier\_\_n\_estimators: 200
                            \item classifier\_\_min\_child\_weight: 5
                            \item classifier\_\_max\_depth: 15
                            \item classifier\_\_learning\_rate: 0.01
                            \item classifier\_\_colsample\_bytree: 0.6
                            \item scaler: passthrough
                        \end{enumerate}
                    \item \textbf{Augmented Data:}
                        \begin{enumerate}
                            \item classifier\_\_subsample: 0.6
                            \item classifier\_\_n\_estimators: 200
                            \item classifier\_\_min\_child\_weight: 1
                            \item classifier\_\_max\_depth: 10
                            \item classifier\_\_learning\_rate: 0.01
                            \item classifier\_\_colsample\_bytree: 0.6
                            \item scaler: passthrough
                        \end{enumerate}
                \end{enumerate}
            \item \textbf{Human Comparison:}
                \begin{enumerate}
                    \item \textbf{Raw Data:}
                        \begin{enumerate}
                            \item classifier\_\_subsample: 0.6
                            \item classifier\_\_n\_estimators: 100
                            \item classifier\_\_min\_child\_weight: 1
                            \item classifier\_\_max\_depth: 15
                            \item classifier\_\_learning\_rate: 0.01
                            \item classifier\_\_colsample\_bytree: 0.8
                            \item scaler: MinMaxScaler()
                        \end{enumerate}
                    \item \textbf{Augmented Data:}
                        \begin{enumerate}
                            \item classifier\_\_subsample: 1.0
                            \item classifier\_\_n\_estimators: 100
                            \item classifier\_\_min\_child\_weight: 1
                            \item classifier\_\_max\_depth: 15
                            \item classifier\_\_learning\_rate: 0.2
                            \item classifier\_\_colsample\_bytree: 0.6
                            \item scaler: StandardScaler()
                        \end{enumerate}
                \end{enumerate}
        \end{enumerate}
    \item \textbf{CLIFS Support Vector Machine:}
        \begin{enumerate}
            \item \textbf{Overall:}
                \begin{enumerate}
                    \item \textbf{Raw Data:}
                        \begin{enumerate}
                            \item classifier\_\_C: 1
                            \item classifier\_\_degree: 2
                            \item classifier\_\_gamma: scale
                            \item classifier\_\_kernel: linear
                            \item scaler: passthrough
                        \end{enumerate}
                    \item \textbf{Augmented Data:}
                        \begin{enumerate}
                            \item classifier\_\_C: 1
                            \item classifier\_\_degree: 2
                            \item classifier\_\_gamma: scale
                            \item classifier\_\_kernel: linear
                            \item scaler: passthrough
                        \end{enumerate}
                \end{enumerate}
            \item \textbf{Human Comparison:}
                \begin{enumerate}
                    \item \textbf{Raw Data:}
                        \begin{enumerate}
                            \item classifier\_\_C: 1
                            \item classifier\_\_degree: 2
                            \item classifier\_\_gamma: scale
                            \item classifier\_\_kernel: linear
                            \item scaler: passthrough
                        \end{enumerate}
                    \item \textbf{Augmented Data:}
                        \begin{enumerate}
                            \item classifier\_\_C: 0.1
                            \item classifier\_\_degree: 6
                            \item classifier\_\_gamma: scale
                            \item classifier\_\_kernel: poly
                            \item scaler: minmax
                        \end{enumerate}
                \end{enumerate}
        \end{enumerate}
\end{enumerate}

Final hyperparameters for all \textbf{regressors} performing \textbf{identity fusion prediction} (all trained on augmented data):
    
\begin{enumerate}
    \item \textbf{CLIFS Random Forest:}
        \begin{enumerate}
            \item \textbf{Overall:}
                \begin{enumerate}
                    \item regressor\_\_max\_depth: 20
                    \item regressor\_\_min\_samples\_leaf: 1
                    \item regressor\_\_min\_samples\_split: 2
                    \item regressor\_\_n\_estimators: 100
                    \item scaler: MinMaxScaler() 
                \end{enumerate}
            \item \textbf{Human Comparison:}
                \begin{enumerate}
                    \item regressor\_\_max\_depth: 20
                    \item regressor\_\_min\_samples\_leaf: 1
                    \item regressor\_\_min\_samples\_split: 2
                    \item regressor\_\_n\_estimators: 200
                    \item scaler: passthrough
                \end{enumerate}
        \end{enumerate}
\end{enumerate}

Final hyperparameters for all \textbf{classifiers} performing \textbf{violence risk prediction}:

\begin{enumerate}
    \item \textbf{VRI with CLIFS:}
        \begin{enumerate}
            \item classifier\_\_max\_depth: None
            \item classifier\_\_min\_samples\_leaf: 2
            \item classifier\_\_min\_samples\_split: 10
            \item classifier\_\_n\_estimators: 100
            \item scaler: passthrough
        \end{enumerate}
    \item \textbf{VRI Random Forest}
        \begin{enumerate}
            \item classifier\_\_max\_depth: None
            \item classifier\_\_min\_samples\_leaf: 2
            \item classifier\_\_min\_samples\_split: 5
            \item classifier\_\_n\_estimators: 300
            \item scaler: StandardScaler()
        \end{enumerate}
\end{enumerate}

\subsection{Compute Resources:}

The resources required to fine-tune the ModernBERT LLM classifier:

\begin{enumerate}
    \item 1x NVIDIA A100 80GB GPU
    \item Time: $\approx$ 1 hour per model hyperparameter search $+$ training
\end{enumerate}

The resources required to train the CLIFS and SBERT random forests and CLIFS SVM classifiers:

\begin{enumerate}
    \item 1x Ryzen 7 9700X CPU
    \item CLIFS RF Time: 0.22--0.39 hours per model hyperparameter search $+$ training (not including the fine-tuning of ModernBERT from above)
    \item SBERT RF Time: $\approx$ CLIFS RF Time
    \item CLIFS SVM Time: 0.06--0.17 hours per model hyperparameter search $+$ training
\end{enumerate}

Next, the resources required to train the CLIFS XGBoost model classifier:

\begin{enumerate}
    \item 1x NVDIA RTX 4070 Ti 12GB GPU
    \item Time: 0.77--1.03 hours per model hyperparameter search $+$ training
\end{enumerate}

Last, the resources required for the regressors:

\begin{enumerate}
    \item 1x Ryzen 7 9700X CPU
    \item $\approx$ 8.3 hours per model hyperparameter search $+$ training
\end{enumerate}

\clearpage

\section{Appendix}
\label{sec:appendixc}

\subsection{Unquestioning Affiliation Index}
\label{sec:appendixuai}

The Unquestioning Affiliation Index is calculated as follows:

\begin{equation}
  \label{eq:UAI}
  \textnormal{UAI} = z(A) - z(C)
\end{equation}

where z-scores ($z(x)=\frac{z-\mu}{\sigma}$; number of standard deviations, $\sigma$, from the mean, $\mu$) standardize counts of affiliation words (A) and cognitive-processing words (C) against the sample distribution.

Our na\"{i}ve UAI which simply removes z-scores and subtracts raw scores:

\begin{equation}
  \label{eq:nUAI}
  \textnormal{nUAI}  = A - C
\end{equation}

\subsection{Violence Risk Index}
\label{sec:appendixvri}

Let \(\overline{A}\) denote the mean of the scores for the four highly significant categories\footnote{Fusion, out-group dehumanization, justification of violence, and explicit calls to and announcements of violence.}, \(\overline{B}\) the mean of the three statistically significant categories\footnote{Out-group slurs, out-group demonization, and hopelessness of alternative solutions.}, and \(\overline{C}\) the mean of the five other relevant categories\footnote{Existential threat, conspiracy belief, inevitable war, martyrdom narrative, and violent role model.}. Then calculate the weighted sum of the means.

\begin{equation}
\label{eq:vri}
\begin{array}{l}
\displaystyle
\overline{A} = \frac{1}{4}\sum_{i=1}^4 A_i,\; \overline{B} = \frac{1}{3}\sum_{j=1}^3 B_j,\; \overline{C} = \frac{1}{5}\sum_{k=1}^5 C_k,\\
\textnormal{VRI} = 100\left(0.54\,\overline{A} + 0.25\,\overline{B} + 0.21\,\overline{C}\right)
\end{array}
\end{equation}

The original VRI assigns ``low,'' ``medium,'' ``high,'' and ``very high'' classifications. We map low and medium to Moderate, high to Ideologically extreme, and very high to Violent self-sacrificial in our analysis. The class thresholds are as follows: VRI $< 10 = \textit{low}$, $10 \geq$ VRI $\leq 30 = \textit{medium}$, $30 <$ VRI $\leq 70 = \textit{high}$, $70 <$ VRI $= \textit{very high}$~\cite{EbnerKavanaghWhitehouse2024}. 

\subsection{Spearman Correlation}
\label{sec:appendixspear}

The Spearman correlation coefficient, $r_s$, is defined as:

\begin{equation}
    r_s = 1 - \frac{6\,\sum_{i=1}^n d_i^2}{n(n^2-1)},
    \qquad
    d_i = R(x_i)-R(y_i).
\end{equation}

where $R(x_i)$ and $R(y_i)$ are the ranks of variables $x_i$ and $y_i$. The difference in ranks is represented by $d_i$ for the $i$-th pair of $x$ and $y$. Spearman correlation measures the monotonic relationship between two variables by comparing the ranked values rather than their raw magnitudes. Direction is indicated by $+$ or $-$.

\end{document}